\documentclass[letterpaper]{article} 
\usepackage[]{aaai2027}  
\usepackage[hyphens]{url}  
\usepackage{graphicx} 
\usepackage{natbib}  
\usepackage{caption} 
\usepackage{algorithm}
\usepackage{algorithmic}

\usepackage{newfloat}
\usepackage{listings}
\DeclareCaptionStyle{ruled}{labelfont=normalfont,labelsep=colon,strut=off} 
\floatstyle{ruled}
\newfloat{listing}{tb}{lst}{}
\floatname{listing}{Listing}

\usepackage{booktabs}
\usepackage{xcolor}  
\usepackage{amsmath}
\usepackage{amsthm}
\usepackage{amssymb}
\usepackage{mathtools}
\usepackage{tcolorbox}
\usepackage{multirow}
\usepackage{subcaption}
\usepackage{colortbl} %
\nocopyright 

\title{Knowledge Restoration-driven Prompt Optimization: Unlocking LLM Potential for Open-Domain Relational Triplet Extraction}
\author{
    Xiaonan Jing\textsuperscript{1,2},
    Gongqing Wu\textsuperscript{1,2}\corresponding,
    Xingrui Zhuo\textsuperscript{1,2}\corresponding,
    Lang Sun\textsuperscript{3},
    Jiapu Wang\textsuperscript{4}
}
\affiliations{
    \textsuperscript{\rm 1}The Key Laboratory of Knowledge Engineering with Big Data (Ministry of Education of China),\\ Hefei University of Technology, China \\
    \textsuperscript{\rm 2}School of Computer Science and Information Engineering, Hefei University of Technology, China \\
    \textsuperscript{\rm 3}Anhui Zhongke Guojin Intelligent Technology Co., Ltd., China \\
    \textsuperscript{\rm 4}Nanjing University of Science and Technology, China \\
    \{jxn, zxr,sl\}@mail.hfut.edu.cn, 
    wugq@hfut.edu.cn, 
    jiapu.wang@njust.edu.cn 

}

\begin{document}

\maketitle

\begin{abstract}
Open-domain Relational Triplet Extraction (ORTE) aims to mine structured knowledge without predefined relation schemas. Large Language Models (LLMs) have advanced ORTE toward a prompt-driven paradigm through powerful in-context learning. However, adapting their extraction behavior to varying open-domain contexts remains challenging. Existing methods typically rely on manually crafted prompts that remain fixed across inputs, despite substantial variation in linguistic expressions and contextual structures. This mismatch may lead to unsupported triplets, while the absence of ground-truth annotations makes such deficiencies difficult to identify and correct. Moreover, free-form relation generation produces non-canonical relation surface forms, undermining knowledge graph consistency.
To address these challenges, we propose Knowledge Restoration-driven Prompt Optimization (\textbf{KRPO}), a framework for label-free target-corpus adaptation. KRPO restores extracted triplets into textual statements and evaluates their semantic consistency with the source inputs, deriving intrinsic feedback without gold annotations. This feedback is transformed into natural-language optimization guidance for batch-wise prompt optimization and adaptation. KRPO further introduces a Memory-augmented Relation Canonicalizer that aligns free-form relations with a dynamically updated schema memory, improving relation consistency. Experiments on three ORTE benchmarks with multiple LLM backbones demonstrate strong overall performance, with KRPO achieving the best average F1 score across the evaluated settings.
\end{abstract}
\begin{links}
\link{Code}{https://anonymous.4open.science/r/KRPO-B26W}
\end{links}


\section{Introduction}
Knowledge Graphs (KGs) have become an essential foundation for modern artificial intelligence~\cite{KG, KG1}, providing structured knowledge for downstream tasks such as information retrieval~\cite{IR}, logical reasoning~\cite{kgreason1}, and question answering systems~\cite{kgqa}.
To construct high-quality KGs, Relational Triplet Extraction (RTE) is a core task that formalizes unstructured text into structured triplets of the form $(subject, relation, object)$.
In practice, traditional RTE methods are constrained by predefined relations and supervised annotations, making them difficult to adapt to the unknown open domain. Therefore, Open-domain Relational Triplet Extraction (ORTE) has emerged. By extracting knowledge without predefined schemas, ORTE serves as a vital infrastructure for understanding open-world knowledge~\cite{openie07}.

The emergence of Large Language Models (LLMs)~\cite{pan2024unifying,wang2024large} has fundamentally reshaped the landscape of ORTE.
Equipped with rich parametric knowledge and strong in-context learning capabilities, LLMs enable prompt-driven ORTE without requiring task-specific fine-tuning on the target corpus~\cite{chatie,llm4ie2}.
With crafted prompts, LLMs have achieved promising performance in ORTE~\cite{EDC}.


\begin{figure}[t]
    \centering
    \includegraphics[width=\linewidth]{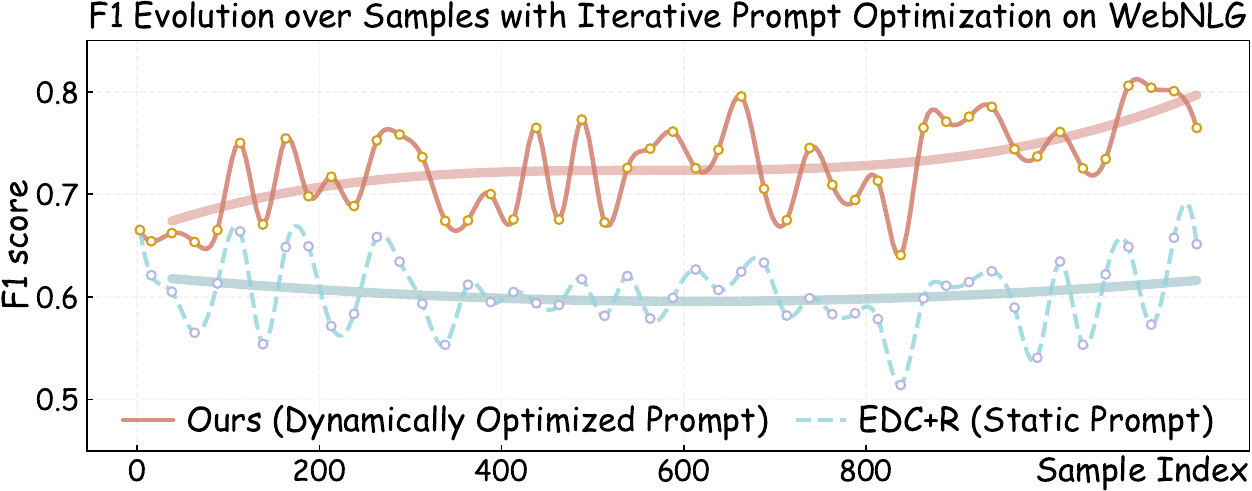}
    \caption{Comparison between KRPO (Ours) and EDC+R~\protect\cite{EDC}. KRPO dynamically optimizes prompts via knowledge restoration-based feedback, unlocking the potential of LLMs for ORTE, whereas EDC+R relies on static prompts and suffers from performance stagnation.}
    \label{fig:intro}
\end{figure}

However, despite these advances, existing LLM-based ORTE methods~\cite{EDC,KGGen} still struggle to effectively unlock the potential of LLMs in open-domain target corpora. 
A fundamental challenge is \textbf{how to adapt LLMs to ORTE across diverse contexts}.
Although LLMs encode extensive relational knowledge in their parameters, whether this knowledge can be elicited in a task-aligned manner depends heavily on the interaction between the task instructions and the input context.
Open-domain texts vary substantially in linguistic expressions and contextual structures, whereas their manually crafted prompts typically remain unchanged across inputs.
This mismatch may drift extraction focus away from source-grounded relations toward generic semantic associations or contextually salient but unsupported information, resulting in omitted relations or hallucinated triplets.
This adaptation challenge is further compounded by the absence of ground-truth annotations, which prevents direct evaluation and correction of extraction behavior on the target corpus.

As illustrated in Figure~\ref{fig:intro}, EDC+R~\cite{EDC} shows limited performance gains as subsequent target samples are processed, since its static prompt cannot adapt to varying inputs. This limitation affects LLM-based ORTE methods~\cite{SAC-KG, KGGen}, which lack a mechanism for optimizing prompts through source-grounded feedback. The mismatch persists throughout inference and constrains adaptation to the target corpus. The key challenge is therefore to derive a source-grounded adaptation signal from the extraction process, enabling prompt optimization to progressively adapt the extraction behavior of LLMs to the target corpus without access to gold annotations.

Beyond extraction quality, ORTE also suffers from \textbf{non-canonical relation surface forms}. 
Since open-domain relations are generated in free-form language, semantically equivalent relations may appear in different surface forms, such as ``\texttt{birthplace}'' and ``\texttt{born in}''. 
Directly retaining these variants leads to redundant relation schemas and undermines consistency of KGs, therefore reducing their practical utility.
Existing canonicalization methods~\cite{CESI,EDC} often rely on coarse semantic matching, which may overlook fine-grained interactions required for relation alignment while allowing schema expansion in open-domain scenarios.

To address this challenge, we propose \textbf{Knowledge Restoration-driven Prompt Optimization} (KRPO), a source-grounded self-reflective adaptation framework for ORTE. KRPO restores extracted triplets into textual statements and evaluates their semantic consistency with the corresponding inputs, thereby deriving intrinsic feedback without ground-truth annotations. This feedback is transformed into textual gradients, namely natural-language optimization guidance, to identify extraction deficiencies and guide prompt rewriting. In this way, the extraction prompt can progressively adapt to the varying contexts of the target corpus.
For improving relation consistency of the resulting KGs, KRPO introduces a Memory-augmented Relation Canonicalizer that aligns free-form relations with dynamic schema memory through fine-grained semantic matching.
As illustrated by the orange track in Figure~\ref{fig:intro}, KRPO replaces static prompting with a batch-wise adaptation, allowing the extraction guidance to evolve with the target corpus, and shows continuous improvement in performance.

In summary, we have made the following contributions:
\begin{itemize}
    \item We propose KRPO, a closed-loop prompt optimization framework for ORTE that enables label-free corpus adaptation by intrinsic feedback from knowledge restoration.
    \item We design a Memory-augmented Relation Canonicalizer to fine-grained align relation expressions with dynamic schema memory, alleviating relation fragmentation.
    \item Experiments on 3 benchmarks with multiple LLM backbones show strong overall performance, with KRPO achieving the best average F1 score.
\end{itemize}

\section{Related Work}
Early Relational Triplet Extraction (RTE) methods rely on predefined schemas and supervised annotations, using either pipeline architectures~\cite{pipline1,Pipeline3} that separate entity and relation extraction~\cite{NER,RE}, or joint models~\cite{OneRel,TPLinker,SpanRE} that mitigate error propagation. Although pre-trained models~\cite{REBEL} improve extraction, they still require domain-specific supervision. Recent work has shifted toward open-domain extraction~\cite{openie07}, using LLMs and multi-agent systems via multi-turn dialogue~\cite{chatie}, code generation~\cite{codeie}, or collaborative frameworks~\cite{KGGen,karma,SAC-KG}. Despite zero- and few-shot flexibility~\cite{AutoKGSurvey,graphrag}, these methods remain sensitive to prompts and generate inconsistent relations, motivating quality-aware extraction frameworks for downstream KG reasoning~\cite{kgreason2,huang2026relink} and completion~\cite{wang2023survey,wang2024ime}.

Traditional prompt optimization~\cite{promptsurvey,promptgpt} uses supervised discrete search~\cite{autoprompt,hardprompt} or continuous prompt tuning~\cite{prefixtuning,prompttuning,ptuning}, limiting interpretability and open-domain applicability. Recent black-box methods optimize prompts using reinforcement learning or LLM optimizers~\cite{RLPrompt, APE, OPRO}, including text-based gradient approximations~\cite{textgrad}, but they rely on task-specific rewards, labeled validation signals, or generic critiques. In contrast, KRPO targets supervision-deficient ORTE by deriving intrinsic feedback from triplet-to-text restoration and source-text consistency checking. It is therefore an ORTE-oriented instantiation of feedback-driven prompt refinement rather than a general prompt optimizer.

Schema-extensible ORTE produces lexically diverse but semantically redundant relations~\cite{openie07}, further worsened by LLM generation~\cite{llm4ie1}. Existing methods use external knowledge~\cite{wordnet}, clustering~\cite{relcanon,CESI,PPDB}, Siamese networks~\cite{releme}, or embedding-based matching~\cite{EDC}, but may over-generalize fine-grained distinctions or rely on shallow similarities. KRPO integrates fine-grained relation-semantic alignment with dynamic schema memory to reduce relation fragmentation, and couples it with prompt optimization to improve KG consistency in the open domain.

\section{Preliminaries}
This section formally defines ORTE, specifies the prompt optimization objective for mining the potential of LLMs, and outlines the problem of relation canonicalization.

\subsection{Open Relational Triplet Extraction}
Given an unstructured corpus $X$, ORTE aims to extract relational triplets for constructing KGs without a predefined relation ontology. For an input text $x \in X$, the goal is to extract $M$ triplets $\mathcal{T}\ =\ \{\left(s_i,\ r_i,\ o_i\right)\}_{\left\{i=1\right\}}^M$,
where $s_i$ and $o_i$ are subject and object in $x$, and $r_i$ is their semantic relation. These triplets represent facts expressed in $x$. Unlike closed-domain extraction with fixed schemas, the relation set $\mathcal{R}_c=\left\{r_c^1,r_c^2,\ldots,r_c^N\right\}$ in open-domain settings is incomplete or empty, and can be dynamically extensible.
\subsection{Prompt LLM-based ORTE}
Recent LLM-based ORTE methods frame the task as conditional generation. Given input text $x$ and prompt $\mathcal{P}$, an LLM $\mathcal{M}$ generates triplets $\mathcal{T}$ by maximizing $\hat{\mathcal{T}}=\arg\max_{\tau}P_{\mathcal{M}}(\mathcal{T}\mid x,\mathcal{P})$.
However, static prompts limit adaptive extraction across diverse contexts, making triplet quality dependent on prompt expressiveness.
To better exploit LLMs in open-domain settings, we formulate a surrogate optimization perspective from $\mathcal{T}$ to $\mathcal{P}$, seeking an improved prompt: $\mathcal{P}^*=\arg\max_{\mathcal{P}}P_{\mathcal{M}}(\mathcal{P}\mid\mathcal{T},x)$.
The optimized prompt is expected to induce high-quality triplet generation for $x$. See Appendix~\ref{validate_prompt_opt} for validation.
\subsection{Semantic-based Relation Canonicalization}
Triplets from the prior stage often contain redundant relations, where distinct phrases express the same meaning. Let $\mathcal{R}_{\mathrm{raw}}=\{r_1,r_2,\ldots,r_n\}$ be the extracted relations, relation canonicalization maps them to canonical schemas $R_c$ through $f:\mathcal{R}_{\mathrm{raw}}\to\mathcal{R}_c$, such that similar relations $r_i$ and $r_c^i$ map to the same canonical form:
\begin{equation}
\small
    f(r_{i})=f\big(r_{c}^{i}\big)\Leftarrow\mathrm{sem}(r_{i})\approx\mathrm{sem}\big(r_{c}^{i}\big),
\end{equation}
where $sem(\cdot)$ denotes semantic meaning. This similarity is measured in a Cross-Encoder semantic space to ensure consistency in the KGs.

\subsection{Task Formulation}
We formulate ORTE as the composition of prompt-guided triplet extraction and relation canonicalization:
\begin{equation}
\small
    \mathcal{KG}^{*}=\text{RelCanon}\{\text{ORTE}[x,\text{P-Opt}(X,\mathcal{P})]\},
\end{equation}
where $\text{P-Opt}(\cdot)$ optimize the prompt $\mathcal{P}$ under unlabeled corpus $X$, $\text{RelCanon}(\cdot)$ canonicalizes relations extracted.

\begin{figure*}[t]
    \centering
    \includegraphics[width=\textwidth]{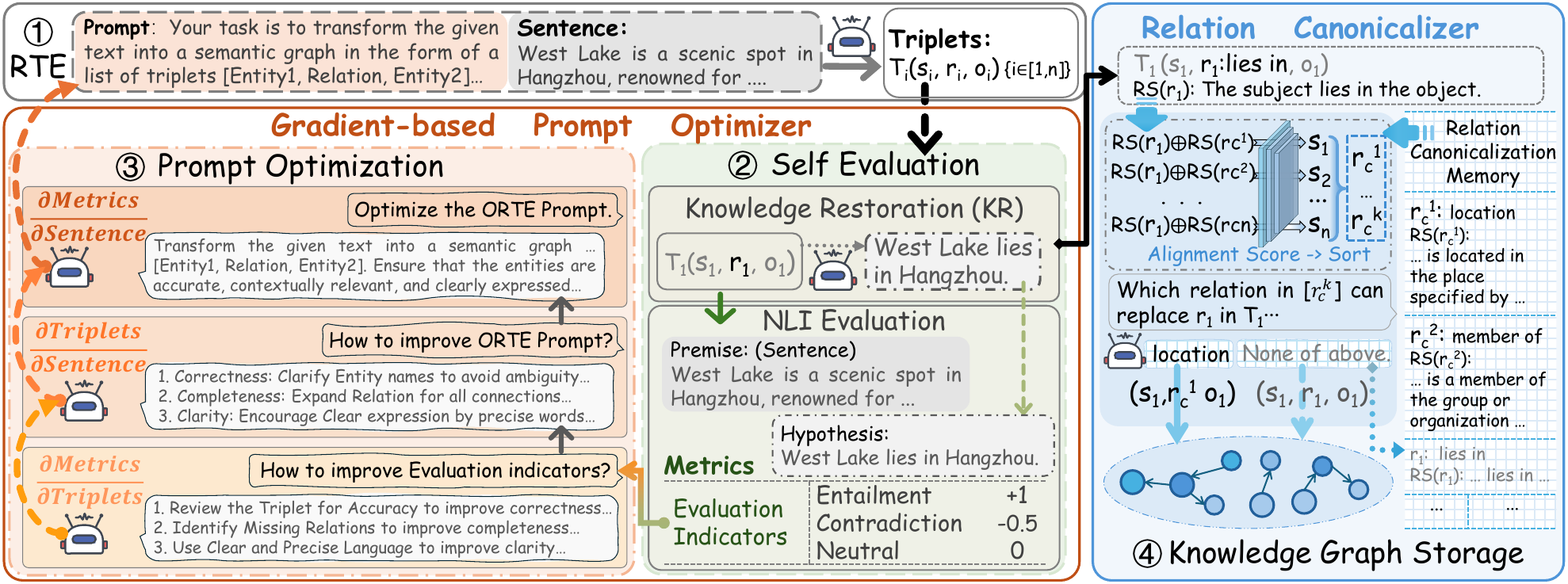}
    \caption{Overview of KRPO. Comprises four modules: (1) Relational Triplet Extraction (RTE), extracting triplets by LLM with an optimizable prompt; (2) Self Evaluation, assessing consistency via NLI on restored text; (3) Prompt Optimization, optimizing the prompt using evaluation feedback as gradients; and (4) Relation Canonicalization, aligning relations with dynamic schemas for Knowledge Graph storage.}
    \label{fig:model}
\end{figure*}
\section{Methodology}
KRPO is designed for target-corpus adaptation in ORTE under a label-free test-time setting, refining the extraction prompt from observed texts and self-evaluation feedback without accessing ground-truth triplet annotations. 
As illustrated in Figure~\ref{fig:model}, KRPO consists of two main stages: 
(1) knowledge restoration-driven prompt optimization, which derives intrinsic feedback to iteratively refine the prompt and extract triplets again; and (2) memory-augmented relation canonicalization, which aligns relations in extracted triplets with a dynamically updated schema memory to improve KG consistency.

\subsection{Textual-Gradient Prompt Optimizer}
This stage aims to optimize the relational triplet extraction prompt. Starting with a task-specific initial prompt, we iteratively optimize it over iterations.
\subsubsection{LLM-based ORTE}
Let $p^{(t)}$ be the prompt at iteration $t$. The LLM extracts candidate relational triplets from text $x$:
\begin{equation}
\small
    \hat{T}=f_{\theta}^{\text{ORTE}}\big(\big[p^{(t)};x\big]\big),
\end{equation}
where $f_{\theta}(\cdot)$ denotes an LLM-based operation and $f_{\theta}^{\text{ORTE}}(\cdot)$ denotes the ORTE function.
Each triplet in $\hat{T}$ is denoted as $\hat{t}=(\hat{s},\hat{r},\hat{o})$, which represents (subject, relation, object).
\subsubsection{Knowledge Restoration-based Self-Evaluation}
We evaluate the faithfulness of extracted triplets to the source text in two steps. 

First, each triplet is restored into a natural language sentence by the knowledge restoration module:
\begin{equation}
\small
    h_{\hat{t}}=\text{KR}(\hat{t})=\mathcal{KR}(\hat{s},\hat{r},\hat{o}),
\end{equation}
where $\mathrm{KR}(\cdot)$ is a constrained verbalizer that preserves only the semantics of the triplet without adding external information. It can be implemented with an LLM (its prompt is provided in Appendix~\ref{prompt_KR}) or an equivalent verbalizer.

Second, an NLI evaluator compares the restored hypothesis $h_{\hat{t}}$ with the source text $x$ as the premise:
\begin{equation}
\small
    E_{\hat{t}}=\mathrm{NLI}(x,h_{\hat{t}})\!\in\!\{\text{entailment, neutral, contradiction}\},
\end{equation}
Anchoring the premise to the source text prevents evaluating triplets solely against generated content. 
We implement the evaluator via LLM prompting (Appendix~\ref{prompt_nli}), while standard NLI models remain applicable. Its output is mapped to a discrete consistency score:
\begin{equation}
\small
   \text{Metrics}(\hat{t})=\begin{cases}1,&E_{\hat{t}}=\text{entailment},\\0,&E_{\hat{t}}=\text{neutral},\\-0.5,&E_{\hat{t}}=\text{contradiction}.\end{cases}
\end{equation}
The scores are aggregated over all extracted triplets,
$\mathrm{Metrics}(\hat{T})=\sum_{i=1}^{M}\mathrm{Metrics}(\hat{t}_i)$,
and used as self-evaluation feedback for prompt optimization.

\subsubsection{Textual-Gradient Prompt Optimization}
Following gradient-inspired prompt optimization~\cite{textgrad}, we define a \emph{textual gradient} as natural-language feedback that indicates a direction for prompt revision, which represents text-level credit assignment over discrete texts rather than numerical differentiation or backpropagation.

After extracting $T (Triplets)$ from $x (Sentence)$, KRPO performs self-evaluation and derives textual feedback from the resulting score $Metrics$. This feedback is regarded as a textual gradient that guides prompt optimization toward:$\arg\max_{\mathcal{P}}P_{\mathcal{M}}(\mathcal{P}\mid\mathcal{T},x)$. 
Formally, the textual gradient is generated as:
\begin{equation}
\small
\begin{split}
    g_{\mathcal{P}} = f^{\mathrm{grad}}_{\theta}(\mathrm{Sentence}, \hat{T}, \mathrm{Metrics}(\hat{T})),\\
    g_{\mathcal{P}}\propto\frac{\partial{Metrics}}{\partial{Sentence}}=\frac{\partial{Metrics}}{\partial{Triplets}}\circ\frac{\partial{Triplets}}{\partial{Sentence}}.
\end{split}
\end{equation}
First, we define instruction $I_1$ to guide the LLM to generate feedback as a textual gradient:
\begin{equation}
\small
    g_i^{(1)}=f_{\theta_1}^{\mathrm{grad}}([\:I_1;{ \hat{T}_i; Metrics(\hat{T}_i) }])\propto\frac{\partial{Metrics}}{\partial{Triplets}}.
\end{equation}
Here, $g_p^{(1)}$ diagnoses deficiencies in correctness, completeness, and clarity. While correctness is primarily assessed by the NLI score, completeness is improved through textual gradients. Instruction $I_2$ combined with $p^{(t)}$ and incorporates these three dimensions into prompt-level guidance, encouraging comprehensive and concise extraction while avoiding under-extraction and redundant generation, yielding the second textual gradient:

\begin{equation}
\small
    g_i^{(2)}=f_{\theta_2}^{\mathrm{grad}}([I_2;p^{(t)};x_i,\hat{T}_i,g_i^{(1)}])\propto\frac{\partial{Triplets}}{\partial{Sentence}}.
\end{equation}
\begin{tcolorbox}[colback=white!5!white, colframe=black!75!black,
left=4pt, right=4pt, top=4pt, bottom=4pt,
title=$I_1$. Prompt for Evaluation-driven Feedback]
\small
    You are an expert evaluator for Open-domain Relational Triplet Extraction (ORTE).\\
    Given the extracted triplets: \{$Triplets$\},
    and their NLI-based evaluation \{$Metrics$\},
    provide concise feedback on ``\textbf{how to improve evaluation indicators}'' to improve its quality in terms of
    \textit{1. correctness, 2. completeness, and 3. clarity.}
\end{tcolorbox}

\begin{tcolorbox}[colback=white!5!white, colframe=black!75!black,
    left=4pt, right=4pt, top=4pt, bottom=4pt,
    title=$I_2$. Prompt for Prompt Optimization Guidance]
    \small
    You are an expert prompt engineer for Open-domain Relational Triplet Extraction (ORTE).\\
    Given the ORTE prompt \{$Prompt$\},
    input sentence \{$Sentence$\},
    extracted triplets \{$Triplets$\},
    and the feedback \{$g_i^{(1)}$\},
    provide concise guidance on ``\textbf{how to improve ORTE prompt}'' to improve extraction quality in terms of
    \textit{1. correctness, 2. completeness, and 3. clarity.}
\end{tcolorbox}
Finally, the guidance is aggregated into a batch buffer $\mathcal{F}$ with $B$ instances, and instruction $I_3$ rewrites the prompt:
\begin{equation}
\small %
p'\!=\!f_{\theta_3}^{\mathrm{Opt}}\left(I_3,p^{(t)},\mathcal{F}^{(t)}\right)\!,
\mathcal{F}^{(t)}\!=\!\left\{\!(x_i,\hat{T}_i,g_i^{(2)})\right\}_{i=1}^{B}.
\end{equation}
Batch-level aggregation could reduce the influence of isolated restoration or NLI deviations, providing a more stable optimization signal for feedback-driven prompt optimization.

The candidate prompt $p'$ is evaluated on the same batch, and $p^{(t+1)}\leftarrow p'$ if its batch-level evaluation score improves; otherwise, the current prompt $p^{(t)}$ is retained.

\begin{tcolorbox}[colback=white!5!white, colframe=black!75!black,
    left=4pt, right=4pt, top=4pt, bottom=4pt,
    title=$I_3$. Prompt for Prompt Optimization]
    \small
    You are improving a structured system prompt used for the Open-domain Relational Triplet Extraction task.\\
    The variable to improve is: \\
    \textless VARIABLE \textgreater \{$Prompt$\} \textless/VARIABLE\textgreater\\
    Given the contextual feedback related to this prompt: 
    \textless CONTEXT\textgreater \{$\mathcal{F}$\} \textless/CONTEXT\textgreater\\
    Your task is to follow the improvement strategies in the \textless CONTEXT\textgreater to \textbf{optimize the ORTE prompt} to make it clearer, more effective, and more concise.
\end{tcolorbox}
\subsection{Relation Canonicalization}

For each extracted triplet $\hat{t}=(\hat{s},r_q,\hat{o})$, we first mask the subject and object in its restored sentence $h_{\hat{t}}$ to obtain a contextual relation representation:
\begin{equation}
\mathcal{RS}(r_q)
=
\mathrm{MaskEntity}(h_{\hat{t}};\hat{s},\hat{o}).
\end{equation}
A pre-fine-tuned XLM-RoBERTa Cross-Encoder~\cite{XLMRoBERTa}, whose parameters remain frozen throughout the KRPO process, scores this representation against each canonical schema $r_c^i \in \mathcal{R}_c$ in the relation canonicalization memory:
\begin{equation}
s_i=A_{\theta}\bigl(\mathcal{RS}(r_q),\mathcal{RS}(r_c^i)\bigr),
\end{equation}
where $A_{\theta}$ is the Cross-Encoder scoring function. The $K$ highest-scoring schemas form the candidate set $\mathcal{R}_c^{(K)}$. Details of the pre-experimental fine-tuning are provided in Appendix~\ref{fine_tune}.

\begin{table*}[t]
\centering
\small
\setlength{\tabcolsep}{1mm} 
\renewcommand{\arraystretch}{0.9} 
\begin{tabular}{@{}llcccccccccccccccc@{}}
\toprule
\multirow{2}{*}{Method}   & \multirow{2}{*}{LLMs}         & \multicolumn{3}{c}{WebNLG}& \multicolumn{3}{c}{REBEL} & \multicolumn{3}{c}{Wiki-NRE} & \multirow{2}{*}{Overall Avg.}\\ \cmidrule(lr){3-5}\cmidrule(lr){6-8}\cmidrule(lr){9-11}
                     &             & Partial & Strict & Exact  & Partial & Strict & Exact  & Partial & Strict & Exact  & \\ \midrule
REGEN                &             & 76.7    & 72.0   & 72.3   & -       & -      & -      & -       & -      & -     & - \\
GenIE                &             & -       & -      & -      & 38.5    & 35.6   & 36.4   & 48.4    & 46.3   & 47.8   & -\\ \midrule
CodeIE & GPT-4o-mini & 62.6     & 55.3    & 57.7  & 43.4     & 37.7   & 40.3  & 51.0      & 49.7    & 50.0   & 49.7 \\
ChatIE & GPT-4o-mini & 58.1     & 53.2    & 54.4  & 45.4     & 40.8   & 42.2  & 51.8      & 49.9    & 50.9   & 49.6 \\
KARMA  & GPT-4o-mini & 40.2     & 32.1    & 35.8  & 36.6     & 31.0   & 33.8  & 35.4      & 28.8    & 33.0   & 34.1 \\
KGGen  & GPT-4o-mini & 48.0     & 37.6    & 41.0  & 36.2     & 30.3   & 32.5  & 31.3      & 26.3    & 29.6   & 34.8 \\ \midrule
\multirow{2}{*}{EDC} & GPT-4o-mini & 70.8    & 63.4   & 66.0   & 49.6    & 43.5   & 46.2   & 61.8    & 60.6   & 60.8   & 58.1\\
                     & Mistral-7B  & 64.1    & 56.1   & 58.7   & 47.2    & 40.4   & 43.4   & 55.0    & 53.9   & 54.1  & 52.5 \\ \midrule
\multirow{5}{*}{EDC+R}&DeepSeek-V3 & 76.9 & 71.9 & 73.4 & 52.8 & 48.3 & 50.0 & 65.7 & 64.9 & 65.1 & 63.2 \\
                       &     GPT-5 & 76.3 & 69.5 & 72.2 & 51.5 & 47.5 & 48.7 & 66.2 & 65.3 & 65.6 & 62.5 \\
                       & GPT-4o-mini& 72.0 & 64.3 & 66.7 & 49.4 & 44.5 & 46.4 & 62.0 & 60.8 & 61.0 & 58.6 \\
                       & Qwen3-32B & 67.3 & 60.5 & 62.5 & 51.7 & 47.2 & 49.1 & 58.6 & 57.1 & 57.7 & 56.8 \\
                       & Mistral-7B& 66.3 & 56.7 & 59.7 & 48.4 & 41.3 & 45.0 & 56.5 & 55.2 & 55.5 & 53.8 \\\midrule                     
\multirow{5}{*}{KRPO}  & DeepSeek-V3 & \underline{77.1}    & \textbf{74.3}   & \textbf{75.2}   & \underline{55.2}    & \textbf{50.5}   & \underline{52.0}   & \textbf{68.0}    & \textbf{67.4}   & \textbf{67.4}  & 65.2\\
                     & GPT-5       & \textbf{77.7}    & \underline{72.0}   & \underline{74.2}   & \textbf{55.8}    & \underline{50.4}   & \textbf{52.3}   & \underline{67.1}    & \underline{65.5}   & \underline{65.8}  & 64.5\\
                     & GPT-4o-mini & 74.4$\pm$0.9    & 69.0$\pm$0.7   & 71.2$\pm$0.6   & 51.9$\pm$1.1    & 45.6$\pm$0.9   & 48.9$\pm$0.8   & 64.2$\pm$1.0    & 61.8$\pm$0.6   & 62.1$\pm$0.6   & 61.0\\
                     & Qwen3-32B   & 74.7$\pm$0.5    & 70.1$\pm$0.4   & 71.8$\pm$0.5   & 51.7$\pm$0.6    & 47.4$\pm$0.7   & 48.8$\pm$0.6   & 60.3$\pm$0.7    & 59.2$\pm$0.5   & 59.3$\pm$0.4   & 60.4\\
                     & Mistral-7B  & 70.6$\pm$0.2    & 65.4$\pm$0.3   & 67.2$\pm$0.3   & 51.7$\pm$0.4    & 44.1$\pm$0.5   & 47.9$\pm$0.7   & 59.4$\pm$0.5    & 58.4$\pm$0.5   & 58.7$\pm$0.5   & 58.2\\
                    &\cellcolor{gray!18} Avg. Gain (↑)&\cellcolor{gray!18} +4.4\% &\cellcolor{gray!18} \textbf{+8.7}\%&\cellcolor{gray!18} +7.5\% &\cellcolor{gray!18} +5.0\%  &\cellcolor{gray!18} +4.0\% &\cellcolor{gray!18} +4.5\% &\cellcolor{gray!18} +3.3\%  &\cellcolor{gray!18} +2.9\%  &\cellcolor{gray!18} +2.8\%&\cellcolor{gray!18} +4.8\% \\\bottomrule
\end{tabular}%
\caption{Comparison with SOTA methods on three datasets in terms of F1 score. The best results are highlighted in \textbf{bold}, and the second-best results are \underline{underlined}. The ``Avg.'' column denotes the average performance across all datasets across all evaluation settings. The ``Avg. Gain (↑)'' row reports the mean relative improvement of KRPO over EDC+R across all LLM backbones.}
\label{main_results}
\end{table*}

An LLM then selects a semantically equivalent candidate or returns $\textit{None}$:
\begin{equation}
r'
=
f_{\theta}^{\mathrm{RC}}
\left(
x,\hat{t},r_q,\mathcal{R}_c^{(K)}\cup\{\textit{None}\}
\right).
\end{equation}
If $r'\neq\mathrm{None}$, the canonicalized triplet is $(\hat{s},r',\hat{o})$. Otherwise, $r_q$ is retained and added to the memory,
$\mathcal{R}_c\leftarrow\mathcal{R}_c\cup\{r_q\}$,
allowing subsequent relations to align with it. The decision prompt is provided in Appendix~\ref{rel_choose}.

The memory could be initialized with available canonical schemas or as an empty set when no schema is provided. According to the above algorithm, the relations extracted by the model also expand the memory.

Overall, KRPO performs batch-wise prompt optimization through restoration-based self-evaluation for triplet extraction, followed by relation canonicalization on the extracted triplets. The complete procedure is summarized in Algorithm~\ref{alg_krpo} in Appendix~\ref{appendix_algo}.

\section{Experiments}
\label{sec:experiments}
In this section, we conduct comprehensive experiments, including comparisons with baselines, ablation studies, and analysis to evaluate the effectiveness and impact of KRPO.

\paragraph{Datasets.}
Following EDC~\cite{EDC}, we use its processed versions of WebNLG~\cite{webnlg2020}, REBEL~\cite{rebeldata}, and Wiki-NRE~\cite{wikinre}, and evaluate KRPO under a label-free test-time adaptation protocol, each sample is extracted twice before and after prompt optimization. Gold annotations are not used for prompt optimization or relation-memory updates, but only for final evaluation. Dataset details are provided in Appendix~\ref{dataset_details}.

\paragraph{Evaluation Metrics.}
We use the official WebNLG evaluation script~\cite{webnlg2020} to report standard token-level Precision, Recall, and F1 under 3 matching settings: Exact (full token match), Partial (partial token overlap), and Strict (full subject-relation-object alignment). We have discussed these surface-level metrics for ORTE in Appendix~\ref{appendix:metric_discussion}.

\paragraph{Baselines.} We evaluate and compare KRPO against several state-of-the-art (SOTA) baselines: traditional trained methods REGEN~\cite{ReGen} and GenIE~\cite{GenIE}, as well as advanced LLM or Agent-based methods, including EDC~\cite{EDC} and its variant EDC+R, CodeIE~\cite{codeie}, ChatIE~\cite{chatie}, KARMA~\cite{karma}, and KGGen~\cite{KGGen}. See Appendix~\ref{baseline_details} for detailed baseline descriptions.

\paragraph{Implementation Details.}
We evaluate KRPO with 5 LLM backbones: locally deployed Mistral-7B and Qwen3-32B on 4 NVIDIA A16 GPUs via vLLM, and API-based GPT-4o-mini, GPT-5, and DeepSeek-V3 (Appendix~\ref{model_details}). All models use a decoding temperature of 0; experiments with GPT-4o-mini and smaller backbones are repeated 3 times, with mean results reported. Relation canonicalization uses a fine-tuned XLM-RoBERTa. We set both the prompt-optimization batch size and canonicalization Top-$K$ to 5 and process samples in their original order. For fair comparison, the initial ORTE prompt, few-shot examples, and relation-memory initialization follow EDC~\cite{EDC}; the examples are disjoint from the evaluated datasets and do not introduce additional schema information beyond the baseline setting.

\subsection{Main Results}

Table~\ref{main_results} shows that KRPO generally achieves stronger F1 performance than the compared baselines, including EDC+R and recent LLM-based methods (e.g., CodeIE, ChatIE, KARMA, KGGen), with the best performance across datasets and evaluation settings. The gains are most pronounced under the Strict protocol, suggesting that KRPO enhances extraction fidelity rather than merely increasing partial surface-level matches. Although a few individual entries are tied with or slightly below the strongest baseline, the overall results demonstrate the effectiveness of KRPO across benchmarks with different linguistic characteristics and relation distributions.

We further analyze the impact of LLM backbones with different model capacities. KRPO generally improves over EDC+R across varying model capacities, with especially clear gains on smaller backbones like Mistral-7B. This indicates that our prompt optimization and relation canonicalization effectively mitigate the reasoning limits of lightweight LLMs, narrowing their performance gap with larger models (see Appendix~\ref{full_main_result} for the complete report).

\subsection{Ablation Studies}
To evaluate individual components, we ablate two key modules using GPT-4o-mini as the backbone, and omitting weaker baselines for clarity.
\subsubsection{Ablation on Prompt Optimization}
\begin{figure}[t]
    \centering
    \includegraphics[width=\columnwidth]{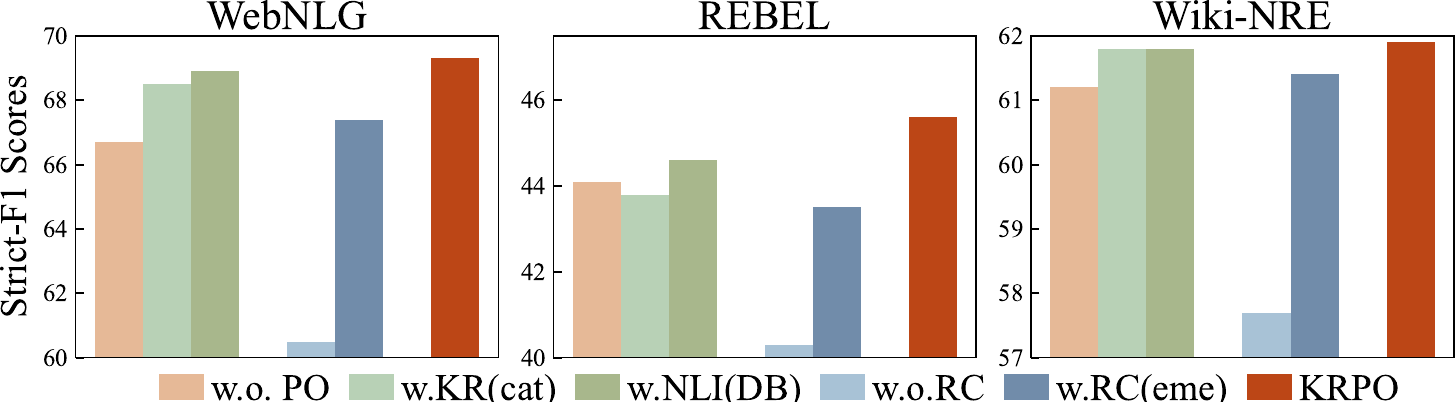}
    \caption{Ablation study on WebNLG, REBEL, and Wiki-NRE. We report the Strict-F1 scores to analyze the contribution of different components (e.g., PO and RC modules) to the final performance.}
    \label{fig:ablation}
\end{figure}

As shown in Figure~\ref{fig:ablation}, removing the prompt optimization module (w.o. PO) consistently degrades performance across datasets, confirming the necessity of iterative optimization. Replacing knowledge restoration (KR) with naive triplet concatenation (w. KR(cat)) yields marginal gains and still trails the full model, indicating that concatenation provides insufficient feedback. Replacing the LLM-based NLI evaluator with DeBERTa-v3-large (w. NLI(DB)) causes a slight performance drop, demonstrating the benefit of stronger semantic feedback from LLMs. Overall, the complete PO module achieves the highest F1 scores, particularly under strict settings, highlighting KR's effectiveness (details in Appendix~\ref{appendix_ablation_PO}).
\subsubsection{Ablation on Relation Canonicalization}
Figure~\ref{fig:ablation} also shows that removing relation canonicalization (w.o. RC) causes substantial performance drops, especially under strict evaluation protocols. Using cosine-similarity-based embedding matching (w. RC(eme)) offers limited recovery, exposing the flaws of shallow semantic similarity. The full RC module performs better, proving its capability to enforce semantic consistency for extracted relations (details in Appendix~\ref{appendix_ablation_RC}).

These ablations suggest that the two components are complementary: prompt optimization mainly improves triplet-level faithfulness, while relation canonicalization improves schema-level consistency.

\begin{figure}[!t]
    \centering
    \includegraphics[width=\columnwidth]{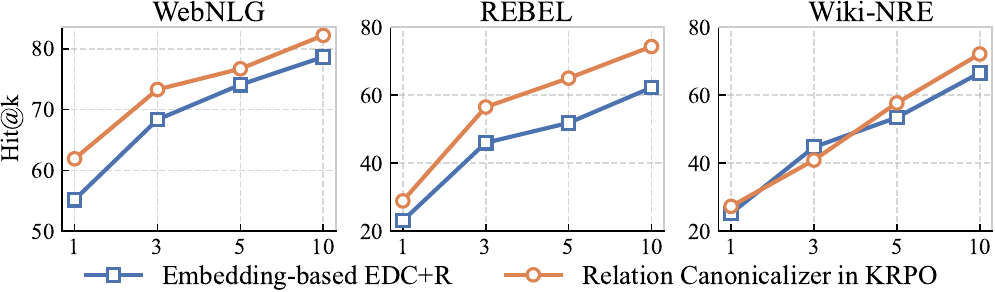}
    \caption{Analysis of Relation Canonicalization Accuracy under different $k$. Evaluated on triplets with correct ``(subject,object)'' pair but incorrect ``relation''.}
    \label{fig:ana_rel_acc}
\end{figure}
\subsection{Analysis}
\begin{figure*}[t]
    \centering
    \includegraphics[width=0.99\textwidth]{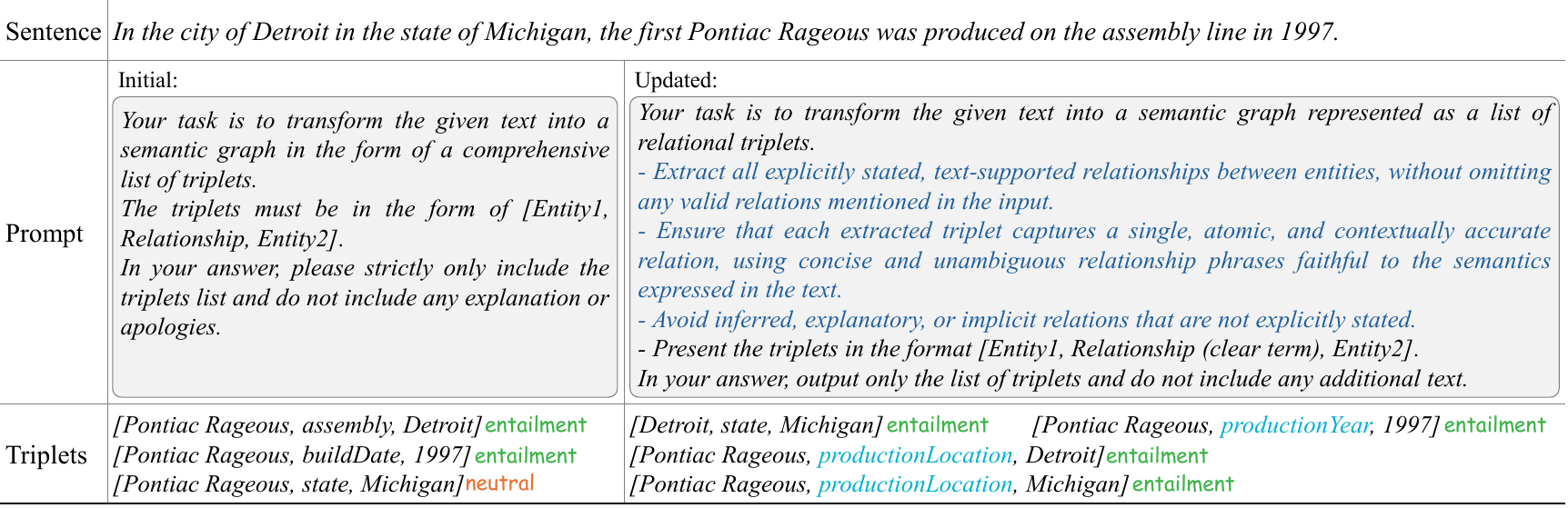}
    \caption{A case study comparing the extraction results between the Initial and Updated prompts. The optimized prompt (right) incorporates more detailed constraints (in blue), guiding the LLM to generate triplets that are more precise compared to the initial prompt (left).}
    \label{fig:case}
\end{figure*}
We further analyze KRPO from several perspectives by focusing on the strongest baselines, EDC and EDC+R. We omit weaker baselines for clarity.
\subsubsection{Analysis of Relation Canonicalization}
\begin{table}[t]
\centering
\small
\begin{tabular}{@{}lcccccc@{}}
\toprule
\multirow{2}{*}{\begin{tabular}[c]{@{}l@{}}Entailment\\ Eval (Partial)\end{tabular}} 
& \multicolumn{2}{c}{WebNLG} 
& \multicolumn{2}{c}{REBEL} 
& \multicolumn{2}{c}{Wiki-NRE} \\ 
\cmidrule(lr){2-3} \cmidrule(lr){4-5} \cmidrule(l){6-7}
& Prop. & F1 & Prop. & F1 & Prop. & F1 \\ 
\midrule
EDC    & 95\% & 68.5 & 96\% & 50.2 & 75\% & 59.7 \\ 
EDC+R  & 95\% & 68.7 & 96\% & 50.5 & 76\% & 60.9 \\
KRPO   & \textbf{98\%} & \textbf{72.7} & \textbf{98\%} & \textbf{52.4} & \textbf{94\%} & \textbf{63.0} \\
\bottomrule
\end{tabular}
\caption{Entailment Triplet Analysis. Best results are in bold.}
\label{tab:entail}
\end{table}
To evaluate relation canonicalization, we analyze samples with correctly predicted entity pairs but misclassified relations. As shown in Figure~\ref{fig:ana_rel_acc}, Cross-Encoder-based scoring consistently outperforms embedding-based similarity across datasets and Hit@$k$ settings, demonstrating stronger fine-grained relation alignment capability (detailed statistics across triplet elements are in Appendix~\ref{appendix_ana_tri_ele}).

\subsubsection{Impact of Self-evaluation}
To evaluate self-evaluation reliability, we analyze entailment triplets under partial matching against static baselines EDC and EDC+R. As shown in Table~\ref{tab:entail}, KRPO retains more entailment triplets and achieves higher F1 scores across datasets, indicating that self-evaluation provides more reliable feedback for extraction optimization (details in Appendix~\ref{appendix:entailment}).

\subsubsection{Effect of Iterative Prompt Optimization}
\begin{figure}[t]
    \centering
    \includegraphics[width=\linewidth]{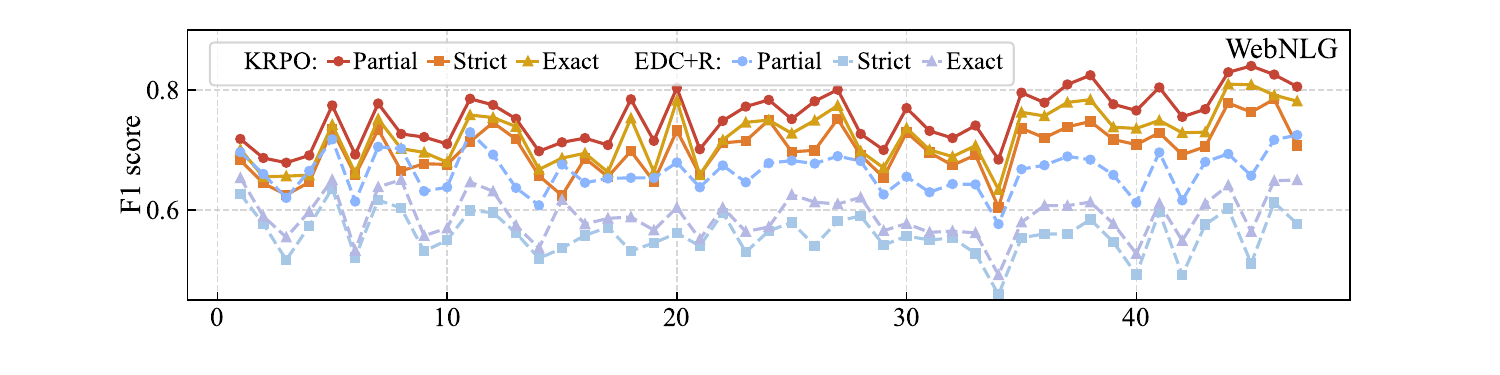}
    \includegraphics[width=\linewidth]{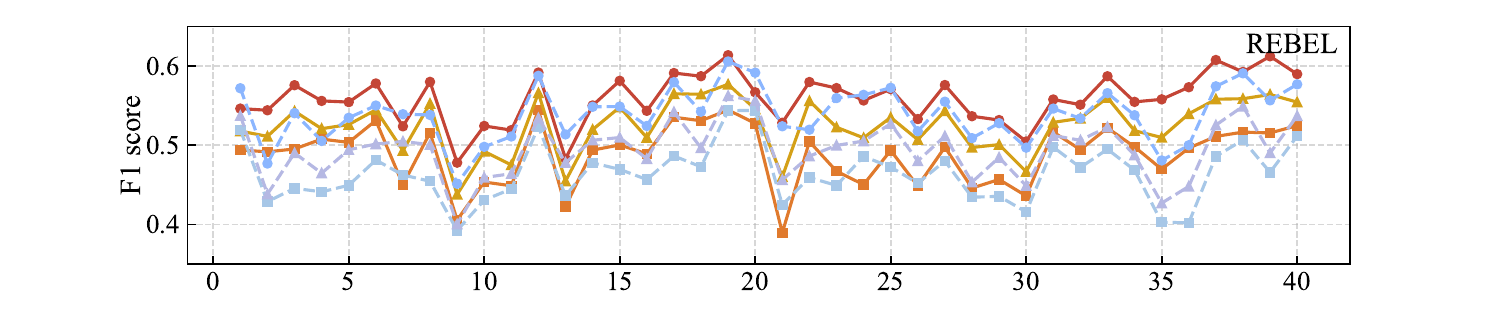}
    \includegraphics[width=\linewidth]{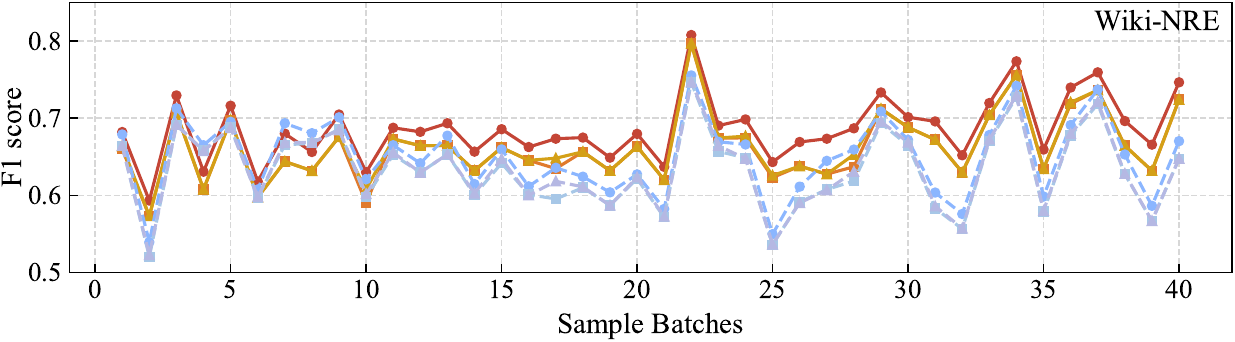}
    \caption{F1 Evolution over Samples with Prompt Optimization on WebNLG, REBEL, and Wiki-NRE datasets.}
    \label{fig:po_datasets}
\end{figure}
To validate iterative prompt optimization, we tracked F1 evolution across consecutive batches of 25 samples, corresponding to 5 consecutive batches with $B=5$. As shown in Figure~\ref{fig:po_datasets}, KRPO consistently improves over static baselines, with larger gains after iterative updates. This demonstrates that our textual gradient-based mechanism effectively accumulates task-level feedback. Notably, gains are most significant under Strict and Exact metrics, indicating that iterative optimization improves not only recall but also structural precision and schema consistency.
During this phase, KRPO incurs an average of 6.3\% per sample higher total-token consumption than EDC+R, while reducing prompt-token usage by 9.3\%, showing a favorable cost-quality trade-off for high-fidelity KG construction. Detailed analysis is provided in Appendix~\ref{sec:token_analysis}.
\subsection{Case Study}
Figure~\ref{fig:case} compares extractions from initial and optimized prompts. For a text describing ``\textit{Pontiac Rageous}'', the initial prompt yields vague, schema-inconsistent triplets and an unclear ``\textit{state}'' relation for ``\textit{Michigan}''. By leveraging self-evaluation feedback, the iteratively optimized prompt guides the LLM toward fine-grained and faithful extractions. Specifically, it correctly isolates the geopolitical relation ``\textit{[Detroit, state, Michigan]}'' and standardizes relations into ``\textit{productionLocation}'' and ``\textit{productionYear}''. This reduces ambiguity and improves KG structural consistency, benefiting downstream tasks like KG reasoning~\cite{kgreason0} (more details in Appendix~\ref{appendix_case_study}).

\section*{Discussion}
\label{sec:limitations}

KRPO has several limitations.  First, its iterative feedback loop requires multiple LLM calls, increasing computation, latency, and token usage.  Although the optimized prompt is reusable, its initial optimization remains costly at scale. 
Smaller restoration and NLI models or model distillation could reduce this cost, although maintaining feedback quality remains challenging.

Second, NLI mainly evaluates the faithfulness of generated triplets and cannot directly detect missing relations. Summed scores may encourage conservative extraction or redundant predictions, while restoration and NLI errors may propagate to prompt updates. Experiments show that using a smaller NLI model that can be deployed results in a small performance degradation. Although constrained restoration, instructions, and batch aggregation mitigate these risks, robustness to evaluation noise and alignment with human judgments remain unverified. Future work should incorporate coverage, redundancy, and confidence into self-evaluation.

Third, optimized prompts may capture corpus-specific patterns, and their trajectories may depend on sample order, batch size, and random seeds.  Cross-dataset transfer and sensitivity to these factors require further evaluation.

Finally, KRPO canonicalizes relations but not entities, potentially creating redundant KG nodes.  The relation memory also scales with schema size.  Future work will study joint entity and relation canonicalization, scalable retrieval, and temporal knowledge modeling.

\section{Conclusion}
In this paper, we analyze the challenges of prompt optimization and relation canonicalization in ORTE and propose KRPO, which enables feedback-driven prompt optimization under supervision-deficient open-domain settings. Through knowledge restoration-based self-evaluation, KRPO iteratively refines prompts, better unlocking the potential of LLMs for extraction. It enhances schema consistency in constructed KGs by memory-augmented relation canonicalization.
Experiments on multiple datasets with different LLM backbones demonstrate the effectiveness of KRPO.

\bibliography{aaai2027}

@inproceedings{EDC,
  author       = {Bowen Zhang and
                  Harold Soh},
  title        = {Extract, Define, Canonicalize: An LLM-based Framework for Knowledge
                  Graph Construction},
  booktitle    = {Empirical Methods in Natural
                  Language Processing},
  pages        = {9820--9836},
  year         = {2024},
  doi          = {10.18653/V1/2024.EMNLP-MAIN.548},
  bibsource    = {dblp computer science bibliography, https://dblp.org}
}

@inproceedings{BGE-M3,
  author       = {Jianlyu Chen and
                  Shitao Xiao and
                  Peitian Zhang and
                  Kun Luo and
                  Defu Lian and
                  Zheng Liu},
  title        = {M3-Embedding: Multi-Linguality, Multi-Functionality, Multi-Granularity
                  Text Embeddings Through Self-Knowledge Distillation},
  booktitle    = {Findings of the Association for Computational Linguistics},
  pages        = {2318--2335},
  year         = {2024},
  doi          = {10.18653/V1/2024.FINDINGS-ACL.137},
  bibsource    = {dblp computer science bibliography, https://dblp.org}
}

@inproceedings{XLMRoBERTa,
  author       = {Alexis Conneau and
                  Kartikay Khandelwal and
                  Naman Goyal and
                  Vishrav Chaudhary and
                  Guillaume Wenzek and
                  Francisco Guzm{\'{a}}n and
                  others},
  title        = {Unsupervised Cross-lingual Representation Learning at Scale},
  booktitle    = {Association for Computational
                  Linguistics},
  pages        = {8440--8451},
  year         = {2020},
  doi          = {10.18653/V1/2020.ACL-MAIN.747},
  bibsource    = {dblp computer science bibliography, https://dblp.org}
}

@inproceedings{GenIE,
  author       = {Martin Josifoski and
                  Nicola De Cao and
                  Maxime Peyrard and
                  Fabio Petroni and
                  Robert West},
  title        = {GenIE: Generative Information Extraction},
  booktitle    = {North American Chapter of
                  the Association for Computational Linguistics},
  pages        = {4626--4643},
  year         = {2022},
  doi          = {10.18653/V1/2022.NAACL-MAIN.342},
  bibsource    = {dblp computer science bibliography, https://dblp.org}
}

@inproceedings{ReGen,
  author       = {Pierre L. Dognin and
                  Inkit Padhi and
                  Igor Melnyk and
                  Payel Das},
  title        = {ReGen: Reinforcement Learning for Text and Knowledge Base Generation
                  using Pretrained Language Models},
  booktitle    = {Empirical Methods in Natural
                  Language Processing},
  pages        = {1084--1099},
  year         = {2021},
  doi          = {10.18653/V1/2021.EMNLP-MAIN.83},
  bibsource    = {dblp computer science bibliography, https://dblp.org}
}

@article{AutoKGSurvey,
  author       = {Yuqi Zhu and
                  Xiaohan Wang and
                  Jing Chen and
                  Shuofei Qiao and
                  Yixin Ou and
                  Yunzhi Yao and
                  Shumin Deng and
                  Huajun Chen and
                  Ningyu Zhang},
  title        = {LLMs for knowledge graph construction and reasoning: recent capabilities
                  and future opportunities},
  journal      = {World Wide Web},
  volume       = {27},
  number       = {5},
  pages        = {58},
  year         = {2024},
  doi          = {10.1007/S11280-024-01297-W},
  bibsource    = {dblp computer science bibliography, https://dblp.org}
}

@inproceedings{SAC-KG,
  author       = {Hanzhu Chen and
                  Xu Shen and
                  Qitan Lv and
                  Jie Wang and
                  Xiaoqi Ni and
                  Jieping Ye},
  title        = {{SAC-KG:} Exploiting Large Language Models as Skilled Automatic Constructors
                  for Domain Knowledge Graph},
  booktitle    = {Association for Computational
                  Linguistics},
  pages        = {4345--4360},
  year         = {2024},
  doi          = {10.18653/V1/2024.ACL-LONG.238},
  bibsource    = {dblp computer science bibliography, https://dblp.org}
}

@misc{graphrag,
      title={From Local to Global: A Graph RAG Approach to Query-Focused Summarization}, 
      author={Darren Edge and Ha Trinh and Newman Cheng and Joshua Bradley and Alex Chao and Apurva Mody and Steven Truitt and Dasha Metropolitansky and Robert Osazuwa Ness and Jonathan Larson},
      year={2025},
      eprint={2404.16130},
      archivePrefix={arXiv},
}

@article{KGGen,
  author       = {Belinda Mo and
                  Kyssen Yu and
                  Joshua Kazdan and
                  Proud Mpala and
                  Lisa Yu and
                  Charilaos. Kanatsoulis and
                  Sanmi Koyejo},
  booktitle = {Neural Information Processing Systems},
  pages = {30092--30115},
  title = {KGGen: Extracting Knowledge Graphs from Plain Text with Language Models},
  volume = {38},
  year = {2025}
}

@inproceedings{openie07,
  author       = {Michele Banko and
                  Michael J. Cafarella and
                  Stephen Soderland and
                  Matthew Broadhead and
                  Oren Etzioni},
  title        = {Open Information Extraction from the Web},
  booktitle    = {International Joint Conference
                  on Artificial Intelligence},
  pages        = {2670--2676},
  year         = {2007},
  bibsource    = {dblp computer science bibliography, https://dblp.org}
}

@inproceedings{autoprompt,
  author       = {Taylor Shin and
                  Yasaman Razeghi and
                  Robert L. Logan IV and
                  Eric Wallace and
                  Sameer Singh},
  title        = {AutoPrompt: Eliciting Knowledge from Language Models with Automatically
                  Generated Prompts},
  booktitle    = {Empirical Methods in Natural
                  Language Processing},
  pages        = {4222--4235},
  year         = {2020},
  doi          = {10.18653/V1/2020.EMNLP-MAIN.346},
  bibsource    = {dblp computer science bibliography, https://dblp.org}
}

@article{pipline1,
  author       = {Dmitry Zelenko and
                  Chinatsu Aone and
                  Anthony Richardella},
  title        = {Kernel Methods for Relation Extraction},
  journal      = {J. Mach. Learn. Res.},
  volume       = {3},
  pages        = {1083--1106},
  year         = {2003},
  bibsource    = {dblp computer science bibliography, https://dblp.org}
}

@inproceedings{Pipeline3,
  author       = {Yee Seng Chan and
                  Dan Roth},
  title        = {Exploiting Syntactico-Semantic Structures for Relation Extraction},
  booktitle    = {Association for Computational Linguistics},
  pages        = {551--560},
  year         = {2011},
  bibsource    = {dblp computer science bibliography, https://dblp.org}
}

@inproceedings{NER,
  title={{Overview of results of the MUC-6 evaluation}},
  author={Sundheim, Beth M},
  booktitle    = {Conference on Message Understanding},
  pages        = {13--31},
  year={1995},
  doi          = {10.3115/1072399.1072402},
}

@inproceedings{RE,
  title={{Overview of MUC-7}},
  author={Chinchor, Nancy},
  booktitle    = {Conference on Message Understanding},
  year={1998},
}

@inproceedings{OneRel,
  author       = {Yuming Shang and
                  Heyan Huang and
                  Xianling Mao},
  title        = {OneRel: Joint Entity and Relation Extraction with One Module in One
                  Step},
  booktitle    = {Association for the Advancement of Artificial Intelligence},
  pages        = {11285--11293},
  year         = {2022},
  doi          = {10.1609/AAAI.V36I10.21379},
  bibsource    = {dblp computer science bibliography, https://dblp.org}
}

@inproceedings{TPLinker,
  author       = {Yucheng Wang and
                  Bowen Yu and
                  Yueyang Zhang and
                  Tingwen Liu and
                  Hongsong Zhu and
                  Limin Sun},
  title        = {TPLinker: Single-stage Joint Extraction of Entities and Relations
                  Through Token Pair Linking},
  booktitle    = {Computational
                  Linguistics},
  pages        = {1572--1582},
  year         = {2020},
  doi          = {10.18653/V1/2020.COLING-MAIN.138},
  bibsource    = {dblp computer science bibliography, https://dblp.org}
}

@inproceedings{SpanRE,
  author       = {Kalpit Dixit and
                  Yaser Al{-}Onaizan},
  title        = {Span-Level Model for Relation Extraction},
  booktitle    = {Association for Computational
                  Linguistics},
  pages        = {5308--5314},
  year         = {2019},
  doi          = {10.18653/V1/P19-1525},
  bibsource    = {dblp computer science bibliography, https://dblp.org}
}

@inproceedings{REBEL,
  author       = {Pere{-}Llu{\'{\i}}s Huguet Cabot and
                  Roberto Navigli},
  title        = {{REBEL:} Relation Extraction By End-to-end Language generation},
  booktitle    = {Findings of the Association for Computational Linguistics: {EMNLP}},
  pages        = {2370--2381},
  year         = {2021},
  doi          = {10.18653/V1/2021.FINDINGS-EMNLP.204},
  bibsource    = {dblp computer science bibliography, https://dblp.org}
}

@misc{chatie,
      title={ChatIE: Zero-Shot Information Extraction via Chatting with ChatGPT}, 
      author={Xiang Wei and Xingyu Cui and Ning Cheng and Xiaobin Wang and Xin Zhang and Shen Huang and Pengjun Xie and Jinan Xu and Yufeng Chen and Meishan Zhang and Yong Jiang and Wenjuan Han},
      year={2024},
      eprint={2302.10205},
      archivePrefix={arXiv},
}

@article{promptsurvey,
  author       = {Pengfei Liu and
                  Weizhe Yuan and
                  Jinlan Fu and
                  Zhengbao Jiang and
                  Hiroaki Hayashi and
                  Graham Neubig},
  title        = {Pre-train, Prompt, and Predict: {A} Systematic Survey of Prompting
                  Methods in Natural Language Processing},
  journal      = {{ACM} Comput. Surv.},
  volume       = {55},
  number       = {9},
  pages        = {195:1--195:35},
  year         = {2023},
  doi          = {10.1145/3560815},
  bibsource    = {dblp computer science bibliography, https://dblp.org}
}

@inproceedings{promptgpt,
  author       = {Tom B. Brown and
                  Benjamin Mann and
                  Nick Ryder and
                  Melanie Subbiah and
                  Jared Kaplan and
                  Prafulla Dhariwal and
                  others},
  title        = {Language Models are Few-Shot Learners},
  booktitle    = {Neural Information Processing Systems},
  year         = {2020},
  bibsource    = {dblp computer science bibliography, https://dblp.org}
}

@article{hardprompt,
  author       = {Zhengbao Jiang and
                  Frank F. Xu and
                  Jun Araki and
                  Graham Neubig},
  title        = {How Can We Know What Language Models Know},
  journal      = {Trans. Assoc. Comput. Linguistics},
  volume       = {8},
  pages        = {423--438},
  year         = {2020},
  doi          = {10.1162/TACL\_A\_00324},
  bibsource    = {dblp computer science bibliography, https://dblp.org}
}

@inproceedings{prefixtuning,
  author       = {Xiang Lisa Li and
                  Percy Liang},
  title        = {Prefix-Tuning: Optimizing Continuous Prompts for Generation},
  booktitle    = {Association for Computational
                  Linguistics},
  pages        = {4582--4597},
  year         = {2021},
  doi          = {10.18653/V1/2021.ACL-LONG.353},
  bibsource    = {dblp computer science bibliography, https://dblp.org}
}

@inproceedings{prompttuning,
  author       = {Brian Lester and
                  Rami Al{-}Rfou and
                  Noah Constant},
  title        = {The Power of Scale for Parameter-Efficient Prompt Tuning},
  booktitle    = {Empirical Methods in Natural
                  Language Processing},
  pages        = {3045--3059},
  year         = {2021},
  doi          = {10.18653/V1/2021.EMNLP-MAIN.243},
  bibsource    = {dblp computer science bibliography, https://dblp.org}
}

@inproceedings{ptuning,
  author       = {Xiao Liu and
                  Kaixuan Ji and
                  Yicheng Fu and
                  Weng Tam and
                  Zhengxiao Du and
                  Zhilin Yang and
                  Jie Tang},
  title        = {P-Tuning: Prompt Tuning Can Be Comparable to Fine-tuning Across Scales
                  and Tasks},
  booktitle    = {Association for Computational
                  Linguistics},
  pages        = {61--68},
  year         = {2022},
  doi          = {10.18653/V1/2022.ACL-SHORT.8},
  bibsource    = {dblp computer science bibliography, https://dblp.org}
}

@article{textgrad,
  author       = {Mert Y{\"{u}}ksekg{\"{o}}n{\"{u}}l and
                  Federico Bianchi and
                  Joseph Boen and
                  Sheng Liu and
                  Pan Lu and
                  Zhi Huang and
                  Carlos Guestrin and
                  James Zou},
  title        = {Optimizing generative {AI} by backpropagating language model feedback},
  journal      = {Nat.},
  volume       = {639},
  number       = {8055},
  pages        = {609--616},
  year         = {2025},
  doi          = {10.1038/S41586-025-08661-4},
  bibsource    = {dblp computer science bibliography, https://dblp.org}
}

@inproceedings{APE,
  author       = {Yongchao Zhou and
                  Andrei Ioan Muresanu and
                  Ziwen Han and
                  Keiran Paster and
                  Silviu Pitis and
                  Harris Chan and
                  Jimmy Ba},
  title        = {Large Language Models are Human-Level Prompt Engineers},
  booktitle    = {International Conference on Learning Representations},
  year         = {2023},
  bibsource    = {dblp computer science bibliography, https://dblp.org}
}

@inproceedings{RLPrompt,
  author       = {Mingkai Deng and
                  Jianyu Wang and
                  Cheng{-}Ping Hsieh and
                  Yihan Wang and
                  Han Guo and
                  Tianmin Shu and
                  others},
  title        = {RLPrompt: Optimizing Discrete Text Prompts with Reinforcement Learning},
  booktitle    = {Empirical Methods in Natural
                  Language Processing},
  pages        = {3369--3391},
  year         = {2022},
  doi          = {10.18653/V1/2022.EMNLP-MAIN.222},
  bibsource    = {dblp computer science bibliography, https://dblp.org}
}

@inproceedings{OPRO,
  author       = {Chengrun Yang and
                  Xuezhi Wang and
                  Yifeng Lu and
                  Hanxiao Liu and
                  Quoc V. Le and
                  Denny Zhou and
                  Xinyun Chen},
  title        = {Large Language Models as Optimizers},
  booktitle    = {International Conference on Learning Representations},
  year         = {2024},
  url          = {https://openreview.net/forum?id=Bb4VGOWELI},
  bibsource    = {dblp computer science bibliography, https://dblp.org}
}

@inproceedings{llm4ie1,
  author       = {Somin Wadhwa and
                  Silvio Amir and
                  Byron C. Wallace},
  title        = {Revisiting Relation Extraction in the era of Large Language Models},
  booktitle    = {Association for Computational
                  Linguistics},
  pages        = {15566--15589},
  year         = {2023},
  doi          = {10.18653/V1/2023.ACL-LONG.868},
  bibsource    = {dblp computer science bibliography, https://dblp.org}
}

@article{llm4ie2,
  author       = {Derong Xu and
                  Wei Chen and
                  Wenjun Peng and
                  Chao Zhang and
                  Tong Xu and
                  Xiangyu Zhao and
                  others},
  title        = {Large language models for generative information extraction: a survey},
  journal      = {Frontiers Comput. Sci.},
  volume       = {18},
  number       = {6},
  pages        = {186357},
  year         = {2024},
  doi          = {10.1007/S11704-024-40555-Y},
  bibsource    = {dblp computer science bibliography, https://dblp.org}
}

@inproceedings{relcanon,
  author       = {Luis Gal{\'{a}}rraga and
                  Geremy Heitz and
                  Kevin Murphy and
                  Fabian M. Suchanek},
  title        = {Canonicalizing Open Knowledge Bases},
  booktitle    = {Conference
                  on Information and Knowledge Management},
  pages        = {1679--1688},
  year         = {2014},
  doi          = {10.1145/2661829.2662073},
  bibsource    = {dblp computer science bibliography, https://dblp.org}
}

@inproceedings{releme,
  author       = {Rifki Afina Putri and
                  Giwon Hong and
                  Sung{-}Hyon Myaeng},
  title        = {Aligning Open {IE} Relations and {KB} Relations using a Siamese Network
                  Based on Word Embedding},
  booktitle    = {International Conference on Computational
                  Semantics},
  pages        = {142--153},
  year         = {2019},
  doi          = {10.18653/V1/W19-0412},
  bibsource    = {dblp computer science bibliography, https://dblp.org}
}

@article{wordnet,
  author       = {George A. Miller},
  title        = {WordNet: {A} Lexical Database for English},
  journal      = {Commun. {ACM}},
  volume       = {38},
  number       = {11},
  pages        = {39--41},
  year         = {1995},
  doi          = {10.1145/219717.219748},
  bibsource    = {dblp computer science bibliography, https://dblp.org}
}

@inproceedings{PPDB,
  author       = {Juri Ganitkevitch and
                  Benjamin Van Durme and
                  Chris Callison{-}Burch},
  title        = {{PPDB:} The Paraphrase Database},
  booktitle    = {North American Chapter
                  of the Association of Computational Linguistics},
  pages        = {758--764},
  year         = {2013},
  bibsource    = {dblp computer science bibliography, https://dblp.org}
}

@inproceedings{CESI,
  author       = {Shikhar Vashishth and
                  Prince Jain and
                  Partha P. Talukdar},
  title        = {{CESI:} Canonicalizing Open Knowledge Bases using Embeddings and Side
                  Information},
  booktitle    = {Conference on World Wide Web},
  pages        = {1317--1327},
  year         = {2018},
  doi          = {10.1145/3178876.3186030},
  bibsource    = {dblp computer science bibliography, https://dblp.org}
}

@book{KG,
  author       = {Aidan Hogan and
                  Eva Blomqvist and
                  Michael Cochez and
                  Claudia d'Amato and
                  Gerard de Melo and
                  Claudio Gutierrez and
                  others},
  title        = {Knowledge Graphs},
  series       = {Synthesis Lectures on Data, Semantics, and Knowledge},
  year         = {2021},
  doi          = {10.2200/S01125ED1V01Y202109DSK022},
  isbn         = {978-3-031-00790-3},
  bibsource    = {dblp computer science bibliography, https://dblp.org}
}

@book{KG1,
  author       = {Jeff Z. Pan and
                  Guido Vetere and
                  Jos{\'{e}} Manu{\'{e}}l G{\'{o}}mez{-}P{\'{e}}rez and
                  Honghan Wu},
  title        = {Exploiting Linked Data and Knowledge Graphs in Large Organisations},
  year         = {2017},
  doi          = {10.1007/978-3-319-45654-6},
  isbn         = {978-3-319-45652-2},
  bibsource    = {dblp computer science bibliography, https://dblp.org}
}

@inproceedings{IR,
  author       = {Chenyan Xiong and
                  Russell Power and
                  Jamie Callan},
  title        = {Explicit Semantic Ranking for Academic Search via Knowledge Graph
                  Embedding},
  booktitle    = {Conference on World Wide Web},
  pages        = {1271--1279},
  year         = {2017},
  doi          = {10.1145/3038912.3052558},
  bibsource    = {dblp computer science bibliography, https://dblp.org}
}

@inproceedings{kgreason0,
  author       = {Xingrui Zhuo and
                  Jiapu Wang and
                  Gongqing Wu and
                  Shirui Pan and
                  Xindong Wu},
  title        = {Effective Instruction Parsing Plugin for Complex Logical Query Answering
                  on Knowledge Graphs},
  booktitle    = {Conference on World Wide Web},
  pages        = {4780--4792},
  year         = {2025},
  doi          = {10.1145/3696410.3714794},
  bibsource    = {dblp computer science bibliography, https://dblp.org}
}

@article{kgreason1,
  author       = {Xingrui Zhuo and
                  Gongqing Wu and
                  Zan Zhang and
                  Xindong Wu},
  title        = {Geometric-Contextual Mutual Infomax Path Aggregation for Relation
                  Reasoning on Knowledge Graph},
  journal      = {{IEEE} Trans. Knowl. Data Eng.},
  volume       = {36},
  number       = {7},
  pages        = {3076--3090},
  year         = {2024},
  doi          = {10.1109/TKDE.2024.3360258},
  bibsource    = {dblp computer science bibliography, https://dblp.org}
}

@inproceedings{kgreason2,
  author       = {Xingrui Zhuo and
                  Shirui Pan and
                  Jiapu Wang and
                  Gongqing Wu and
                  Zan Zhang and
                  Rui Li and
                  Zizhong Wei and
                  Xindong Wu},
  title        = {Progressive Prefix-Memory Tuning for Complex Logical Query Answering
                  on Knowledge Graphs},
  booktitle    = {International Joint Conference on
                  Artificial Intelligence},
  pages        = {3716--3724},
  year         = {2025},
  doi          = {10.24963/IJCAI.2025/413},
  bibsource    = {dblp computer science bibliography, https://dblp.org}
}

@inproceedings{kgqa,
  author       = {Yuyu Zhang and
                  Hanjun Dai and
                  Zornitsa Kozareva and
                  Alexander J. Smola and
                  Le Song},
  title        = {Variational Reasoning for Question Answering With Knowledge Graph},
  booktitle    = {Association for the Advancement of Artificial Intelligence},
  pages        = {6069--6076},
  year         = {2018},
  doi          = {10.1609/AAAI.V32I1.12057},
  bibsource    = {dblp computer science bibliography, https://dblp.org}
}

@inproceedings{webnlg2020,
  title={The 2020 bilingual, bi-directional webnlg+ shared task overview and evaluation results (webnlg+ 2020)},
  author={Ferreira, Thiago Castro and Gardent, Claire and Ilinykh, Nikolai and Van Der Lee, Chris and Mille, Simon and Moussallem, Diego and Shimorina, Anastasia},
  booktitle={International Workshop on Natural Language Generation from the Semantic Web},
  year={2020}
}

@inproceedings{rebeldata,
  author       = {Pere{-}Llu{\'{\i}}s Huguet Cabot and
                  Roberto Navigli},
  title        = {{REBEL:} Relation Extraction By End-to-end Language generation},
  booktitle    = {Findings of the Association for Computational Linguistics: {EMNLP}},
  pages        = {2370--2381},
  year         = {2021},
  doi          = {10.18653/V1/2021.FINDINGS-EMNLP.204},
  bibsource    = {dblp computer science bibliography, https://dblp.org}
}

@inproceedings{wikinre,
  author       = {Bayu Distiawan Trisedya and
                  Gerhard Weikum and
                  Jianzhong Qi and
                  Rui Zhang},
  title        = {Neural Relation Extraction for Knowledge Base Enrichment},
  booktitle    = {Association for Computational
                  Linguistics},
  pages        = {229--240},
  year         = {2019},
  doi          = {10.18653/V1/P19-1023},
  bibsource    = {dblp computer science bibliography, https://dblp.org}
}

@inproceedings{TEKGEN,
  author       = {Oshin Agarwal and
                  Heming Ge and
                  Siamak Shakeri and
                  Rami Al{-}Rfou},
  title        = {Knowledge Graph Based Synthetic Corpus Generation for Knowledge-Enhanced
                  Language Model Pre-training},
  booktitle    = {North American Chapter of
                  the Association for Computational Linguistics},
  pages        = {3554--3565},
  year         = {2021},
  doi          = {10.18653/V1/2021.NAACL-MAIN.278},
  bibsource    = {dblp computer science bibliography, https://dblp.org}
}

@inproceedings{codeie,
  author       = {Peng Li and
                  Tianxiang Sun and
                  Qiong Tang and
                  Hang Yan and
                  Yuanbin Wu and
                  Xuanjing Huang and
                  Xipeng Qiu},
  title        = {CodeIE: Large Code Generation Models are Better Few-Shot Information
                  Extractors},
  booktitle    = {Association for Computational
                  Linguistics},
  pages        = {15339--15353},
  year         = {2023},
  doi          = {10.18653/V1/2023.ACL-LONG.855},
  bibsource    = {dblp computer science bibliography, https://dblp.org}
}

@inproceedings{karma,
  title     = {KARMA: Leveraging Multi-Agent LLMs for Automated Knowledge Graph Enrichment},
  author    = {Yuxing Lu and Wei Wu and Xukai Zhao and Rui Peng and Jinzhuo Wang},
  booktitle = {Neural Information Processing Systems},
  year      = {2025},
}

@article{pan2024unifying,
  title={Unifying large language models and knowledge graphs: A roadmap},
  author={Shirui Pan and Linhao Luo and Yufei Wang and Chen Chen and Jiapu Wang and Xindong Wu},
  journal={{IEEE} Trans. Knowl. Data Eng.},
  volume={36},
  number={7},
  pages={3580--3599},
  year={2024},
}

@inproceedings{wang2024large,
  author       = {Jiapu Wang and
                  Kai Sun and
                  Linhao Luo and
                  Wei Wei and
                  Yongli Hu and
                  Alan Wee{-}Chung Liew and
                  Shirui Pan and
                  Baocai Yin},
  title        = {Large Language Models-guided Dynamic Adaptation for Temporal Knowledge
                  Graph Reasoning},
  booktitle    = {Neural Information Processing Systems},
  year         = {2024},
  bibsource    = {dblp computer science bibliography, https://dblp.org}
}

@article{wang2023survey,
  title={A survey on temporal knowledge graph completion: Taxonomy, progress, and prospects},
  author={Jiapu Wang and Boyue Wang and Meikang Qiu and Shirui Pan and Bo Xiong and Heng Liu and Linhao Luo and Tengfei Liu and Yongli Hu and Baocai Yin and others},
  journal      = {CoRR},
  volume       = {abs/2308.02457},
  year={2023}
}

@inproceedings{wang2024ime,
  title={IME: Integrating multi-curvature shared and specific embedding for temporal knowledge graph completion},
  author={Jiapu Wang and Zheng Cui and Boyue Wang and Shirui Pan and Junbin Gao and Baocai Yin and Wen Gao},
  booktitle    = {Conference on World Wide Web},
  pages={1954--1962},
  year={2024}
}

@inproceedings{huang2026relink,
  author       = {Manzong Huang and
                  Chenyang Bu and
                  Yi He and
                  Xingrui Zhuo and
                  Xindong Wu},
  title        = {Relink: Constructing Query-Driven Evidence Graph On-the-Fly for GraphRAG},
  booktitle    = {Association for the Advancement of Artificial Intelligence},
  pages        = {31202--31210},
  year         = {2026},
  doi          = {10.1609/AAAI.V40I37.40382}
}
\clearpage
\appendix


\section{The Use of Large Language Models}
We used a large language model (LLM) solely as an editing assistant to improve the grammar, clarity,and concision of the manuscript. All technical contributions, experimental design, data processing.evaluation, and conclusions reported in the paper were authored and verified by the human authors.LLM-suggested edits were reviewed and accepted or modified by the authors; no numerical results, figures, or analyses were generated or approved solely by the LLM.
\section{Ethics statement}
We confirm that our work does not involve human subjects, sensitive personal data, or experiments that may cause harm to individuals or groups. The datasets used are publicly available and no personally identifiable information is included. Our methodology and findings are intended for academic purposes and do not pose foreseeable risks of misuse. We have carefully considered issues of fairness, bias, and privacy, and to the best of our knowledge, our research maintains integrity and complies with all applicable ethical standards.
\section{Method Details}
\subsection{The KRPO Algorithm}
\label{appendix_algo}

We provide the comprehensive algorithmic procedure of our proposed KRPO framework in Algorithm~\ref{alg_krpo}. For each batch, KRPO first performs prompt optimization with label-free acceptance and then canonicalizes the accepted triplets before processing the next batch.

\begin{algorithm}[]
    \caption{KRPO: Knowledge Restoration-driven Prompt Optimization}
    \label{alg_krpo}
    \textbf{Input}: Dataset $\mathcal{X}$, Initial Prompt $p$,
    Relation Memory $\mathcal{R}_c$\\
    \textbf{Parameter}: Batch Size $B$, Instructions $I_1, I_2, I_3$\\
    \textbf{Output}: Final Accepted Prompt $p^*$,
    Knowledge Graph $\mathcal{G}$

    \begin{algorithmic}[1]
        \STATE Initialize $\mathcal{G} \leftarrow \emptyset$.

        \FOR{each batch $\mathcal{B} \subset \mathcal{X}$
        with $|\mathcal{B}|=B$}

            \STATE \textbf{Phase 1: Textual-Gradient Prompt Optimization}
            \STATE Initialize feedback buffer
            $\mathcal{F} \leftarrow \emptyset$ and
            batch score $S_{\mathcal{B}} \leftarrow 0$.

            \FOR{each sentence $x \in \mathcal{B}$}
                \STATE $\hat{T}_x \leftarrow \text{LLM}(x,p)$
                \hfill \textit{// First extraction}

                \STATE $S_x \leftarrow
                \sum_{\hat{t}\in\hat{T}_x}
                \text{NLI}(x,\text{Restore}(\hat{t}))$
                \hfill \textit{// Label-free self-evaluation}

                \STATE $S_{\mathcal{B}}
                \leftarrow S_{\mathcal{B}} + S_x$

                \STATE $g_p^{(1)} \leftarrow
                \text{LLM}(I_1,\hat{T}_x,S_x)$
                \hfill \textit{// Evaluation feedback}

                \STATE $g_p^{(2)} \leftarrow
                \text{LLM}(I_2,p,x,\hat{T}_x,g_p^{(1)})$
                \hfill \textit{// Optimization guidance}

                \STATE Append $(x,g_p^{(2)})$ to $\mathcal{F}$.
            \ENDFOR

            \STATE $p' \leftarrow \text{LLM}(I_3,p,\mathcal{F})$
            \hfill \textit{// Candidate prompt}

            \STATE Initialize candidate batch score
            $S'_{\mathcal{B}} \leftarrow 0$.

            \FOR{each sentence $x \in \mathcal{B}$}
                \STATE $\hat{T}'_x \leftarrow \text{LLM}(x,p')$
                \hfill \textit{// Second extraction}

                \STATE $S'_x \leftarrow
                \sum_{\hat{t}\in\hat{T}'_x}
                \text{NLI}(x,\text{Restore}(\hat{t}))$

                \STATE $S'_{\mathcal{B}}
                \leftarrow S'_{\mathcal{B}} + S'_x$
            \ENDFOR

            \IF{$S'_{\mathcal{B}} > S_{\mathcal{B}}$}
                \STATE $p \leftarrow p'$
                \FOR{each sentence $x \in \mathcal{B}$}
                    \STATE $\hat{T}^{acc}_x \leftarrow \hat{T}'_x$
                \ENDFOR
            \ELSE
                \FOR{each sentence $x \in \mathcal{B}$}
                    \STATE $\hat{T}^{acc}_x \leftarrow \hat{T}_x$
                \ENDFOR
            \ENDIF

            \STATE \textbf{Phase 2: Relation Canonicalization}

            \FOR{each sentence $x \in \mathcal{B}$}
                \FOR{each triplet
                $\hat{t}=(\hat{s},r_q,\hat{o})
                \in\hat{T}^{acc}_x$}

                    \STATE $h_{\hat{t}}
                    \leftarrow
                    \text{MaskEntity}(\text{Restore}(\hat{t}))$

                    \STATE $\mathcal{R}_{topK}
                    \leftarrow
                    \text{CrossEncoder}
                    (h_{\hat{t}},\mathcal{R}_c)$

                    \STATE $r'
                    \leftarrow
                    \text{LLM-Decision}
                    (r_q,\mathcal{R}_{topK})$

                    \IF{$r'$ is $None$}
                        \STATE $\mathcal{R}_c
                        \leftarrow
                        \mathcal{R}_c\cup\{r_q\}$
                        \hfill \textit{// Dynamic Schema Expansion}
                        \STATE $r' \leftarrow r_q$
                    \ENDIF

                    \STATE Add $(\hat{s},r',\hat{o})$
                    to $\mathcal{G}$.
                \ENDFOR
            \ENDFOR
        \ENDFOR

        \STATE Set $p^* \leftarrow p$.
        \STATE \textbf{return} $p^*,\mathcal{G}$
    \end{algorithmic}
\end{algorithm}

For each batch $\mathcal{B}$, KRPO sequentially performs prompt optimization and relation canonicalization. In Phase 1, the current prompt $p$ is first used to extract triplets $\hat{T}_x$ from each sentence. Knowledge restoration and NLI-based self-evaluation produce the sample-level score $S_x$, while Instructions $I_1$ and $I_2$ generate textual-gradient feedback. The feedback from the $B$ samples is aggregated into $\mathcal{F}$, based on which Instruction $I_3$ generates a candidate prompt $p'$.

The candidate prompt is evaluated through a second extraction pass on the same batch. Let $S_{\mathcal{B}}$ and $S'_{\mathcal{B}}$ denote the aggregate self-evaluation scores obtained using $p$ and $p'$, respectively. If $S'_{\mathcal{B}} > S_{\mathcal{B}}$, the candidate prompt is accepted and $\hat{T}'_x$ is retained as $\hat{T}^{acc}_x$; otherwise, the current prompt and its initial extraction $\hat{T}_x$ are retained. Therefore, each sample undergoes exactly two ORTE extraction passes, and no gold annotations are used for prompt generation or prompt acceptance.

In Phase 2, the accepted triplets $\hat{T}^{acc}_x$ from the current batch undergo relation canonicalization before the next batch is processed. A Cross-Encoder retrieves candidate relations from the current relation memory $\mathcal{R}_c$, and an LLM decision maker selects the appropriate canonical relation. If no suitable candidate exists, the raw relation is retained and added to $\mathcal{R}_c$. Consequently, both the accepted prompt $p$ and the updated relation memory $\mathcal{R}_c$ are carried forward to the next batch, while the canonicalized triplets are incrementally added to the knowledge graph $\mathcal{G}$.

\subsection{Optimization Objective Transfer Validation}
\label{validate_prompt_opt}
In this section, we validate the feasibility of transferring the optimization objective from the output space to the input space through a Bayesian perspective.
Specifically, according to Bayes' theorem, for arbitrary prompt $\mathcal{P}$, triplet $\mathcal{T}$, and input text $x$, we have
\begin{equation}
\normalsize
    P_{\mathcal{M}}(\mathcal{P}\mid\mathcal{T},x)=\frac{P_{\mathcal{M}}(\mathcal{T}\mid x,\mathcal{P})P_{\mathcal{M}}(\mathcal{P}\mid x)}{P_{\mathcal{M}}(\mathcal{T}\mid x)},
\end{equation}
rearranging terms yields the triplet generation probability
\begin{equation}
\normalsize
    P_{\mathcal{M}}(\mathcal{T}\mid x,\mathcal{P})=\frac{P_{\mathcal{M}}(\mathcal{P}\mid\mathcal{T},x)P_{\mathcal{M}}(\mathcal{T}\mid x)}{P_{\mathcal{M}}(\mathcal{P}\mid x)},
\end{equation}
since $P_{\mathcal{M}}(\mathcal{T}\mid x)$ is independent of the prompt $\mathcal{P}$, optimizing the original objective is equivalent to $\arg\max_{\mathcal{P}}P_{\mathcal{M}}(\mathcal{T}\mid\mathcal{P},x)<\arg\max_{\mathcal{P}}\frac{P_{\mathcal{M}}(\mathcal{P}\mid\mathcal{T},x)}{P_{\mathcal{M}}(\mathcal{P}\mid x)}$. Taking the logarithm, we obtain $\arg\max_{\mathcal{P}}\log P_{\mathcal{M}}(\mathcal{T}\mid\mathcal{P},x)=\arg\max_{\mathcal{P}}\left[\log P_{\mathcal{M}}(\mathcal{P}\mid\mathcal{T},x)-\log P_{\mathcal{M}}\left(\mathcal{P}\mid x\right)\right]$. 

Furthermore, in our feedback-driven prompt optimization process, prompt updates are dominated by task-level feedback induced by the generated triplets. Consequently, the optimization dynamics are primarily driven by the gradient $g_{\mathcal{P}}\log P_{\mathcal{M}}\left(\mathcal{P}\mid\mathcal{T},x\right)$, while the language prior $\log P_{\mathcal{M}}(\mathcal{P}\mid x)$ only serves as a weak regularizer that maintains linguistic fluency and structural validity. Under this assumption, the following approximation holds:
\begin{equation}
\normalsize
    \arg\max_{\mathcal{P}}P_{\mathcal{M}}(\mathcal{T}\mid\mathcal{P},x)\approx\arg\max_{\mathcal{P}}P_{\mathcal{M}}(\mathcal{P}\mid\mathcal{T},x).
\end{equation}

Therefore, by optimizing the reverse probability $P_{\mathcal{M}}(\mathcal{T}\mid x)$ , we can effectively substitute the optimization of the original objective $P_{\mathcal{M}}(\mathcal{T}\mid x)$. 
This reformulation enables us to build higher-quality KGs by optimizing the input space to unlock the potential of LLMs.

Strictly speaking, maximizing $P_{\mathcal{M}}(\mathcal{P}\mid\mathcal{T},x)$ is not equivalent to directly maximizing $P_{\mathcal{M}}(\mathcal{T}\mid\mathcal{P},x)$, since Bayes decomposition introduces a prompt prior term P(P|x). Therefore, we regard this formulation as a surrogate optimization perspective rather than an exact probabilistic equivalence. The motivation is that textual feedback derived from extracted triplets provides an interpretable signal for rewriting prompts toward behaviors associated with higher-quality extraction.

The above derivation provides an intuitive motivation for shifting optimization from output generation to prompt revision. Since prompts and triplets are discrete texts, KRPO does not compute exact gradients over prompt parameters; instead, it uses LLM-generated textual feedback as a gradient-inspired approximation for iterative prompt rewriting.

\subsection{Prompt for Knowledge Restoration}
\label{prompt_KR}
In this section, we introduce a prompt designed for knowledge restoration. The task is to convert a given factual triplet, consisting of a subject, relation, and object, into a natural language sentence. The prompt ensures that no information not present in the triplet is added, while retaining the semantics of the relation. The generated sentence should accurately reflect the meaning of the triplet, maintaining the original context and factual integrity.
\begin{tcolorbox}[colback=white!5!white, colframe=black!75!black, title=Prompt for Knowledge Restoration]
    You are given a factual triplet in the form (subject, relation, object).\\
    Your task is to convert the triplet into a natural language sentence, ensuring the following:\\
    - Do not add any information not present in the triplet.\\
    - Retain the semantics of the relation.\\
    - The sentence should accurately reflect the meaning of the triplet.\\
    - Just need to give one sentence.
\end{tcolorbox}

\subsection{Prompt for NLI}
\label{prompt_nli}
In this section, we present a prompt designed for Natural Language Inference (NLI). The task is to determine the logical relation between a given premise (a long text) and a hypothesis (a short text). The prompt provides three possible labels for the relation: `entailment', where the hypothesis logically follows from the premise; `contradiction', where the hypothesis is logically inconsistent with the premise; and `neutral', where the hypothesis is neither entailed nor contradicted by the premise. The result is returned in a structured JSON format, including the label, confidence score, and reasoning behind the decision. Other NLP tools with NLI capabilities can also be used to evaluate text relations when LLM is not available.
\begin{tcolorbox}[colback=white!5!white, colframe=black!75!black, title=Prompt for NLI]
You are an expert in Natural Language Inference (NLI). Your task is to determine the logical relation between a given premise (a long text) and a hypothesis (a short text).\\
There are three possible labels:\\
1. ``entailment'' - The hypothesis logically follows from the premise.\\
2. ``contradiction'' - The hypothesis is logically inconsistent with the premise.\\
3. ``neutral'' - The hypothesis is neither entailed nor contradicted by the premise.\\
Return the result in the following JSON format:\\
\{\\  \hspace*{1em}``label'': ``entailment'' / ``contradiction'' / ``neutral'',\\
    \hspace*{1em}``confidence'': float between 0 and 1,\\
    \hspace*{1em}``reasoning'': a brief explanation of your decision.\\
\}\\
Premise:\\
Hypothesis:
\end{tcolorbox}

\subsection{Prompt for Relation Canonicalization Decision}
\label{rel_choose}
In this section, we present a prompt designed for the Relation Canonicalization Decision task. The goal of this task is to determine whether a raw relation extracted from a given text can be replaced by an existing canonical relation schema. Given an input text and its extracted relational triplet, the prompt explicitly provides the definition of the query relation and a set of candidate canonical relations with their corresponding schemas. The model is required to assess semantic equivalence and select the most appropriate canonical relation to replace the query relation. If none of the candidates are suitable, the model is expected to indicate that no replacement is applicable. This prompt helps fine-grained semantic alignment of relations and supports the dynamic expansion of the canonical relation set.
\begin{tcolorbox}[colback=white!5!white, colframe=black!75!black, title=Prompt for Relation Canonicalization Decision]
    Given the following text and a relational triplet extracted from it:\\
    Text: \\
    Triplet: \\
    The relation ``\textit{query relation}'' in the triplet is defined as ``\textit{query relation schema}'' In this context, is there any one relation appropriate to replace it?\\
    Choices:
    \textit{Relation choices with schemas.}\\
    Output:
\end{tcolorbox}

\section{Model details}
\subsection{Fine-tuning Details of the Relation Canonicalizer}
\label{fine_tune}
We construct the training data from the TEKGEN~\cite{TEKGEN} dataset by filtering samples whose token length is less than 20, resulting in approximately 199,621 instances. Each instance consists of a short text pair $(S_t,S_x)$, where $S_t$ denotes a linearized triplet sequence and $S_x$ is the corresponding natural language sentence.

To construct negative samples, we adopt a sentence replacement strategy. For each positive pair $(S_x,S_t,y=1)$, we randomly sample another instance whose triplet relation differs from that of $S_t$, and use its sentence as a negative sentence $S_x$. This yields a negative pair $(S_x^-,S_t,y=0)$, where $r(S_x^-)\neq r(S_t)$. The number of negative samples is set to be equal to that of positive samples.

Our objective is to learn a semantic consistency scoring function $A_\theta(S_t,S_x)\in\mathbb{R}$. Considering the directional nature of Relation Canonicalization, we concatenate the input sequences and feed them into an XLM-RoBERTa encoder to obtain the joint representation based on the $[CLS]$ token:
\begin{equation}
    h=Encoder_{\text{XLM-RoBERTa}}\left([S_t\oplus S_x\right]),
\end{equation}
where $\oplus$ denotes concatenation using the $[SEP]$ token. 

A linear layer is then applied to compute the matching score $s=W^\top h+b$. Following the training strategy of M3~\cite{BGE-M3}, we introduce a binary cross-entropy loss to guide optimization $\mathcal{L}_{\mathrm{BCE}}=-\left[y\log\sigma(s)+(1-y)\log\left(1-\sigma(s)\right)\right]$, where $\sigma(\cdot)$ denotes the sigmoid function.

In addition, for positive and negative samples with scores $s^+=A_\theta(S_t,S_x),\quad s^-=A_\theta(S_t,S_x^-)$, we introduce a ranking loss to explicitly enlarge the relative score margin between positive and negative pairs: $\mathcal{L}_\mathrm{rank}=\max\left(0,m-(s^+-s^-)\right)$, where $m>0$ is a margin hyperparameter. The final training objective is defined as $\mathcal{L}=\mathcal{L}_{\mathrm{BCE}}+\mathcal{L}_{\mathrm{rank}}$.

\paragraph{Role of the Fine-tuning Stage.}
We emphasize that this fine-tuning stage is an offline auxiliary procedure whose sole purpose is to adapt the Cross-Encoder to relation-semantic alignment. It is not jointly trained with KRPO and is independent of the knowledge restoration, self-evaluation, prompt optimization, and dynamic memory update processes. Once fine-tuned on TEKGEN, the Cross-Encoder remains frozen throughout all KRPO experiments and is used only to retrieve semantically relevant candidate relations for the subsequent LLM-based canonicalization decision. No target-corpus annotations, gold triplets, or feedback generated during KRPO are used to update its parameters. Therefore, the fine-tuning procedure should be regarded as task-specific initialization of the relation retriever and is orthogonal to the online KRPO optimization process.

\subsection{LLMs Details}
\label{model_details}
To ensure fair comparison, we evaluated KRPO based on different Large Language Models (LLMs) from the baseline. This includes a range of models of varying sizes and architectures to evaluate the robustness of our approach on different LLM backbones. The temperature parameter was set to 0 across experiments. For each run, we independently restart the whole KRPO optimization process, including extraction, knowledge restoration, evaluation, and prompt updating. Although the LLM temperature is set to 0, minor variations may arise from nondeterministic serving and iterative optimization trajectories. The specific model we used in our experiment is as follows:
\begin{itemize}
    \item Mistral-7B: \url{https://huggingface.co/mistralai/Mistral-7B-Instruct-v0.3}
    \item Qwen3-32B: \url{https://huggingface.co/Qwen/Qwen3-32B}
    \item GPT-4o-mini: \url{https://platform.openai.com/docs/models/gpt-4o-mini}
    \item GPT-5: \url{https://platform.openai.com/docs/models/gpt-5}
    \item DeepSeek-V3: \url{https://api-docs.deepseek.com/api/deepseek-api}
\end{itemize}

For the NLI evaluator ablation, we use the following encoder-based model as a replacement for the LLM-based evaluator:
\begin{itemize}
    \item DeBERTa-v3-large: \url{https://huggingface.co/microsoft/deberta-v3-large}
\end{itemize}

\section{Experimental Details}
\label{sec:exp_details}
\subsection{Dataset Details}
\label{dataset_details}
To comprehensively evaluate the performance of KRPO in open-domain scenarios, we employ three distinct benchmarks: WebNLG~\cite{webnlg2020}, REBEL~\cite{rebeldata}, and Wiki-NRE~\cite{wikinre}. 
Crucially, we utilize the versions processed by EDC~\cite{EDC}.
The statistics of these processed datasets are summarized in Table~\ref{tab:dataset_stats}.

\paragraph{WebNLG.}
Originally designed for Natural Language Generation (NLG), the WebNLG+2020 (v3.0) dataset maps DBpedia triplets to text. Following the EDC setting, we use a refined subset containing \textbf{1,165 instances} that cover \textbf{159 unique relations}. This dataset serves as a gold-standard benchmark for precise extraction capabilities due to its high-quality human annotations.

\paragraph{REBEL.}
The REBEL dataset is a large-scale silver-standard corpus derived from Wikipedia via a BART-based extraction pipeline. To ensure experimental consistency, we adopt the EDC-processed version, which consists of \textbf{1,000 sampled instances} (filtered from the original 105k+ entries). This subset encompasses \textbf{200 distinct relations}, providing a diverse and semantically complex testbed.

\paragraph{Wiki-NRE.}
Derived from Wikipedia for neural relation extraction, Wiki-NRE is widely used for evaluating relational generalization. Similarly, we utilize the version processed by EDC, which contains \textbf{1,000 instances} sampled from the original corpus. With \textbf{45 relation types}, it evaluates the model's stability on standard relational distributions.

\begin{table}[h]
    \centering
    \small
    \setlength{\tabcolsep}{2pt}
    \begin{tabular}{lcccc}
        \toprule
        Dataset & Source & Original Size & Used Size & \# Relations \\
        \midrule
        WebNLG & DBpedia & - & 1,165 & 159 \\
        REBEL   & Wikipedia & 105,516 & 1,000 & 200 \\
        Wiki-NRE      & Wikipedia & 29,619  & 1,000 & 45  \\
        \bottomrule
    \end{tabular}
    \caption{Statistics of the experimental datasets. All datasets are processed using the EDC framework~\protect\cite{EDC} to ensure high quality and consistency. ``Original Size'' refers to the raw corpus size, while ``Used Size'' indicates the refined subset used in this work.}
    \label{tab:dataset_stats}
\end{table}

\subsection{Baselines Details}
\label{baseline_details}
We compare KRPO against three representative state-of-the-art methods, encompassing both fully supervised models and LLM-based frameworks.

REGEN~\cite{ReGen} (Reinforcement Learning for Text-to-Graph Generation) represents the state-of-the-art among fully supervised approaches on the WebNLG dataset. It is a sequence-to-sequence framework built upon pre-trained language models (such as T5 or BART). A key innovation of REGEN is its integration of Reinforcement Learning (RL) to address the exposure bias problem in standard teacher forcing. By optimizing directly for non-differentiable graph evaluation metrics (e.g., SPICE, METEOR) during training, it achieves robust bidirectional performance for both text-to-graph and graph-to-text tasks.

GenIE~\cite{GenIE} (Generative Information Extraction) is a fully supervised autoregressive formulation that achieves SOTA performance on large-scale relation extraction benchmarks like REBEL and Wiki-NRE. Unlike traditional classification-based methods, GenIE treats extraction as a sequence generation task. Its core contribution is a constrained decoding mechanism (trie-based beam search), which forces the model to generate only valid entity and relation identifiers defined in the schema. This ensures structural validity and schema compliance, making it a strong upper bound for supervised closed-domain extraction.

CodeIE~\cite{codeie} addresses the structural ambiguity inherent in natural language outputs by reformulating information extraction as a code generation task. Instead of prompting Large Language Models (LLMs) to generate standard text, CodeIE utilizes code-style templates (such as Python data classes or functions) to define the extraction schema. This approach leverages the rigorous structural constraints of programming languages and the strong code-generation capabilities of modern LLMs, thereby improving the structural fidelity of the extracted entities and relations.

ChatIE~\cite{chatie} is a zero-shot information extraction framework that transforms the traditional extraction pipeline into a multi-turn conversational question-answering (QA) task. By interacting with an instruction-following LLM through a systematic two-stage dialogue—first identifying potential element types within a sentence, and subsequently querying for specific structural information based on those identified types—ChatIE effectively decomposes complex extraction goals into simpler, manageable sub-tasks without requiring task-specific fine-tuning.

KGGen~\cite{KGGen} is a recent LLM-powered framework aimed at extracting dense and semantically coherent knowledge graphs directly from plain text. While it utilizes LLMs for initial triplet extraction, its primary contribution lies in a novel iterative entity resolution and clustering algorithm. This posterior process dynamically consolidates synonyms and merges semantically equivalent entities, directly mitigating the node sparsity and semantic redundancy issues that frequently plague standard open information extraction methods.

KARMA~\cite{karma} is a multi-agent framework designed for the automated enrichment of knowledge graphs from unstructured text. Moving beyond single-LLM paradigms, KARMA employs a collaborative network of specialized agents assigned to a variety of distinct roles, including entity discovery, relation extraction, and schema alignment, and so on. A key advantage of this framework is its built-in conflict resolution mechanism, which utilizes multi-agent debate and multi-layer verification to cross-check extracted facts, thereby significantly reducing contradictory edges and maintaining high extraction accuracy.

EDC~\cite{EDC} (Extract-Define-Canonicalize) serves as the strongest baseline for open-domain extraction without fine-tuning, reporting SOTA results across all three datasets (WebNLG, REBEL, Wiki-NRE) in zero-shot/few-shot settings. It operates in a three-stage pipeline: (1) \textit{Extract}: prompting a frozen LLM to generate open information extraction (OpenIE) triplets; (2) \textit{Define}: dynamically defining relation candidates; and (3) \textit{Canonicalize}: utilizing an embedding-based retriever to map extracted raw relations to canonical schemas via semantic similarity. 

To further enhance extraction performance, the original authors also introduced \textbf{EDC+R} in EDC~\cite{EDC}, an augmented variant incorporating a posterior \textbf{Refinement} step. Specifically, EDC+R independently performs LLM-based entity recognition and relation generation on the source text, feeding these decoupled structural elements back to the LLM as supplementary context to optimize the initial triplets. In our experiments, we comprehensively evaluate both EDC and EDC+R across diverse LLMs to establish a robust baseline comparison against our proposed approach. As a direct predecessor to our work, it represents the current best practice for LLM-driven knowledge graph construction.

\subsection{Full version of the main experiment}
\label{full_main_result}
\begin{table*}[]
    \renewcommand{\arraystretch}{0.9}
\centering
\begin{tabular}{@{}llccccccccc@{}}
\toprule
                     &             & \multicolumn{9}{c}{WebNLG}                                                            \\ \midrule
Method               & LLM         & \multicolumn{3}{c}{Partial} & \multicolumn{3}{c}{Strict} & \multicolumn{3}{c}{Exact}  \\ \cmidrule(l){3-11} 
&             & Precision & Recall & F1 & Precision & Recall & F1 & Precision & Recall & F1  \\ \midrule
REGEN                &             & 75.5 & \underline{78.8} & 76.7 & 71.3 & 73.5 & \underline{72.0} & 71.4 & 73.8 & 72.3  \\
GenIE                &             & - & - & - & - & - & - & - & - & -  \\ \midrule
CodeIE                 & GPT-4o-mini & 61.0 & 64.3 & 62.6 & 54.2 & 56.4 & 55.3 & 56.6 & 58.8 & 57.7  \\
ChatIE                 & GPT-4o-mini & 57.6 & 58.6 & 58.1 & 52.9 & 53.5 & 53.2 & 54.0 & 54.8 & 54.4  \\
KARMA                  & GPT-4o-mini & 39.3 & 41.1 & 40.2 & 31.7 & 32.5 & 32.1 & 35.3 & 36.3 & 35.8  \\
KGGen                  & GPT-4o-mini & 46.6 & 49.5 & 48.0 & 36.8 & 38.4 & 37.6 & 40.1 & 41.9 & 41.0  \\ \midrule
\multirow{5}{*}{EDC}   & DeepSeek-V3 & 75.2 & 78.3 & 76.7 & 69.1 & 70.9 & 70.0 & 71.8 & 73.6 & 72.7  \\
                       & GPT-5       & 73.0 & 76.7 & 74.8 & 68.0 & 70.2 & 69.1 & 69.7 & 72.1 & 70.9  \\
                       & GPT-4o-mini & 68.9 & 72.8 & 70.8 & 62.3 & 64.5 & 63.4 & 64.6 & 67.5 & 66.0  \\
                       & Qwen3-32B   & 64.7 & 69.9 & 67.2 & 57.6 & 60.9 & 59.2 & 59.8 & 63.1 & 61.4  \\
                       & Mistral-7B  & 61.7 & 66.7 & 64.1 & 54.5 & 57.8 & 56.1 & 57.1 & 60.4 & 58.7  \\ \midrule
\multirow{5}{*}{EDC+R} & DeepSeek-V3 & 74.0 & \textbf{80.0} & 76.9 & 69.7 & \underline{74.2} & 71.9 & 71.2 & \textbf{75.7} & 73.4  \\
                       & GPT-5       & 75.1 & 77.5 & 76.3 & 68.1 & 71.0 & 69.5 & 71.1 & 73.3 & 72.2  \\
                       & GPT-4o-mini & 68.8 & 75.5 & 72.0 & 61.9 & 66.9 & 64.3 & 64.4 & 69.2 & 66.7  \\
                       & Qwen3-32B   & 64.3 & 70.6 & 67.3 & 58.3 & 62.9 & 60.5 & 60.2 & 65.0 & 62.5  \\
                       & Mistral-7B  & 63.5 & 69.4 & 66.3 & 54.9 & 58.6 & 56.7 & 57.8 & 61.7 & 59.7  \\ \midrule
\multirow{5}{*}{KRPO}  & DeepSeek-V3 & \textbf{77.0} & 77.2 & \underline{77.1} & \textbf{72.5} & \textbf{76.2} & \textbf{74.3} & \textbf{75.2} & \underline{75.2} & \textbf{75.2}  \\
                       & GPT-5       & \underline{76.7} & 78.7 & \textbf{77.7} & \underline{71.4} & 72.6 & \underline{72.0} & \underline{73.5} & 74.9 & \underline{74.2}  \\
                       & GPT-4o-mini & 73.7 & 75.1 & 74.4 & 68.6 & 69.4 & 69.0 & 70.8 & 71.6 & 71.2  \\
                       & Qwen3-32B   & 74.2 & 75.2 & 74.7 & 69.7 & 70.5 & 70.1 & 71.5 & 72.1 & 71.8  \\
                       & Mistral-7B  & 69.9 & 71.3 & 70.6 & 64.9 & 65.9 & 65.4 & 66.7 & 67.7 & 67.2  \\ \bottomrule
\end{tabular}%
\caption{Comparison with baseline methods on WebNLG. The best results are highlighted in \textbf{bold}, and the second-best results are \underline{underlined}}
\label{main_results_full}
\end{table*}

\begin{table*}[]
    \renewcommand{\arraystretch}{0.9}
\centering
\begin{tabular}{@{}llccccccccc@{}}
\toprule
                     &              & \multicolumn{9}{c}{REBEL}                                                             \\ \midrule
Method               & LLM          & \multicolumn{3}{c}{Partial} & \multicolumn{3}{c}{Strict} & \multicolumn{3}{c}{Exact}  \\ \cmidrule(l){3-11} 
&              & Precision       & Recall  & F1      & Precision & Recall  & F1     & Precision & Recall & F1      \\ \midrule
REGEN                &              & -       & -       & -       & -       & -       & -      & -       & -      & -       \\
GenIE                &              & 38.1    & 39.1    & 38.5    & 35.3    & 36.1    & 35.6   & 36.2    & 36.9   & 36.4    \\ \midrule
CodeIE                 & GPT-4o-mini & 42.6 & 44.2 & 43.4 & 37.2 & 38.2 & 37.7 & 39.8 & 40.8 & 40.3 \\
ChatIE                 & GPT-4o-mini & 44.8 & 46.0 & 45.4 & 40.5 & 41.1 & 40.8 & 41.8 & 42.6 & 42.2 \\
KARMA                  & GPT-4o-mini & 35.8 & 37.4 & 36.6 & 30.5 & 31.5 & 31.0 & 33.3 & 34.3 & 33.8 \\
KGGen                  & GPT-4o-mini & 35.2 & 37.3 & 36.2 & 29.7 & 30.9 & 30.3 & 31.9 & 33.1 & 32.5 \\ \midrule
\multirow{5}{*}{EDC}   & DeepSeek-V3 & 48.4 & 53.0 & 50.6 & 46.2 & 46.6 & 46.4 & 46.0 & 50.2 & 48.0 \\
                       & GPT-5       & 49.3 & 50.1 & 49.7 & 45.3 & 45.7 & 45.5 & 44.9 & 49.1 & 46.9 \\
                       & GPT-4o-mini & 47.0 & 52.5 & 49.6 & 41.6 & 45.6 & 43.5 & 45.9 & 46.5 & 46.2 \\
                       & Qwen3-32B   & 48.0 & 52.6 & 50.2 & 45.3 & 45.9 & 45.6 & 45.2 & 49.8 & 47.4 \\
                       & Mistral-7B  & 44.7 & 50.0 & 47.2 & 42.0 & 38.9 & 40.4 & 41.3 & 45.7 & 43.4 \\ \midrule
\multirow{5}{*}{EDC+R} & DeepSeek-V3 & 52.2 & 53.4 & 52.8 & 47.9 & 48.7 & 48.3 & 49.6 & 50.4 & 50.0 \\
                       & GPT-5       & 50.7 & 52.3 & 51.5 & 47.0 & 48.0 & 47.5 & 48.2 & 49.2 & 48.7 \\
                       & GPT-4o-mini & 48.5 & 50.3 & 49.4 & 43.9 & 45.1 & 44.5 & 45.7 & 47.1 & 46.4 \\
                       & Qwen3-32B   & 51.0 & 52.4 & 51.7 & 46.9 & 47.5 & 47.2 & 48.6 & 49.6 & 49.1 \\
                       & Mistral-7B  & 47.5 & 49.3 & 48.4 & 40.7 & 41.9 & 41.3 & 44.4 & 45.6 & 45.0 \\ \midrule
\multirow{5}{*}{KRPO}  & DeepSeek-V3 & \underline{54.7} & \underline{55.7} & \underline{55.2} & \underline{50.2} & \textbf{50.8} & \textbf{50.5} & \underline{51.6} & \textbf{52.4} & \underline{52.0} \\
                       & GPT-5       & \textbf{55.3} & \textbf{56.3} & \textbf{55.8} & \textbf{51.9} & \underline{49.0} & \underline{50.4} & \textbf{54.1} & 50.6 & \textbf{52.3} \\
                       & GPT-4o-mini & 51.4 & 52.4 & 51.9 & 45.3 & 45.9 & 45.6 & 46.7 & \underline{51.3} & 48.9 \\
                       & Qwen3-32B   & 50.2 & 53.3 & 51.7 & 47.2 & 47.6 & 47.4 & 50.3 & 47.4 & 48.8 \\
                       & Mistral-7B  & 49.0 & 54.7 & 51.7 & 41.7 & 46.8 & 44.1 & 45.9 & 50.1 & 47.9 \\ \bottomrule
\end{tabular}%
\caption{Comparison with baseline methods on REBEL. The best results are highlighted in \textbf{bold}, and the second-best results are \underline{underlined}}
\label{main_results_full_rebel}
\end{table*}

The experimental results in Tables~\ref{main_results_full}, \ref{main_results_full_rebel}, and \ref{main_results_full_wiki} show that KRPO consistently outperforms all baselines across the three datasets in terms of Precision, Recall, and F1-score. This trend holds under both the ``Strict'' and ``Exact'' settings, where KRPO achieves the best overall F1 scores and exhibits more pronounced gains than in relaxed matching protocols.

We also include four recent baselines based on LLM or Agent, namely CodeIE, ChatIE, KARMA, and KGGen. Although these methods adopt different prompting or generation strategies, they remain weaker than the original EDC and EDC+R baselines, indicating that performance is limited by the underlying extraction formulation with static prompts. In contrast, KRPO consistently yields the strongest results across all metrics and datasets, demonstrating its effectiveness as a more robust structured extraction framework.

In addition, KRPO shows particularly clear advantages on smaller backbones such as Mistral-7B, suggesting that the proposed method is especially effective in improving extraction quality when model capacity is limited. Overall, these results confirm the robustness and adaptability of KRPO across datasets, evaluation settings, and model sizes.

\begin{table*}[]
    \renewcommand{\arraystretch}{0.9}
\centering
\begin{tabular}{@{}llccccccccc@{}}
\toprule
                     &              & \multicolumn{9}{c}{Wiki-NRE}                                                         \\ \midrule
Method               & LLM          & \multicolumn{3}{c}{Partial} & \multicolumn{3}{c}{Strict} & \multicolumn{3}{c}{Exact} \\ \cmidrule(l){3-11} 
&              & Precision       & Recall  & F1      & Precision & Recall  & F1     & Precision & Recall & F1     \\ \midrule
REGEN                &              & -       & -       & -       & -       & -       & -      & -       & -      & -      \\
GenIE                &              & 48.2    & 48.6    & 48.4    & 46.2    & 46.4    & 46.3   & 47.7    & 47.9   & 47.8   \\ \midrule
CodeIE                 & GPT-4o-mini & 50.4 & 51.6 & 51.0 & 49.2 & 50.2 & 49.7 & 49.5 & 50.5 & 50.0 \\
ChatIE                 & GPT-4o-mini & 51.5 & 52.1 & 51.8 & 49.8 & 50.0 & 49.9 & 50.7 & 51.1 & 50.9 \\
KARMA                  & GPT-4o-mini & 33.4 & 37.7 & 35.4 & 27.3 & 30.5 & 28.8 & 31.5 & 34.7 & 33.0 \\
KGGen                  & GPT-4o-mini & 29.8 & 33.0 & 31.3 & 25.2 & 27.5 & 26.3 & 28.5 & 30.8 & 29.6 \\ \midrule
\multirow{5}{*}{EDC}   & DeepSeek-V3 & 63.8 & \underline{67.3} & 65.5 & \underline{66.3} & 63.0 & 64.6 & 64.9 & 64.9 & 64.9 \\
                       & GPT-5       & 65.8 & 66.0 & 65.9 & 63.5 & \underline{67.0} & 65.2 & 65.4 & 65.4 & 65.4 \\
                       & GPT-4o-mini & 61.5 & 62.1 & 61.8 & 60.4 & 60.8 & 60.6 & 59.2 & 62.5 & 60.8 \\
                       & Qwen3-32B   & 57.0 & 62.0 & 59.4 & 55.9 & 60.9 & 58.3 & 57.0 & 60.3 & 58.6 \\
                       & Mistral-7B  & 52.7 & 57.5 & 55.0 & 52.4 & 55.5 & 53.9 & 51.8 & 56.6 & 54.1 \\ \midrule
\multirow{5}{*}{EDC+R} & DeepSeek-V3 & 65.2 & 66.2 & 65.7 & 64.7 & 65.1 & 64.9 & 64.8 & 65.4 & 65.1 \\
                       & GPT-5       & 66.2 & 66.2 & 66.2 & 65.2 & 65.4 & 65.3 & \underline{66.3} & 64.9 & 65.6 \\
                       & GPT-4o-mini & 61.5 & 62.5 & 62.0 & 60.3 & 61.3 & 60.8 & 60.6 & 61.4 & 61.0 \\
                       & Qwen3-32B   & 58.1 & 59.1 & 58.6 & 56.7 & 57.5 & 57.1 & 57.3 & 58.1 & 57.7 \\
                       & Mistral-7B  & 56.0 & 57.0 & 56.5 & 54.8 & 55.6 & 55.2 & 55.1 & 55.9 & 55.5 \\ \midrule
\multirow{5}{*}{KRPO}  & DeepSeek-V3 & \underline{66.3} & \textbf{69.8} & \textbf{68.0} & \textbf{67.3} & \textbf{67.5} & \textbf{67.4} & \textbf{67.3} & \textbf{67.5} & \textbf{67.4} \\
                       & GPT-5       & \textbf{66.9} & \underline{67.3} & \underline{67.1} & 65.4 & 65.6 & \underline{65.5} & 65.8 & \underline{65.8} & \underline{65.8} \\
                       & GPT-4o-mini & 64.2 & 64.2 & 64.2 & 63.5 & 60.2 & 61.8 & 61.9 & 62.3 & 62.1 \\
                       & Qwen3-32B   & 60.3 & 60.3 & 60.3 & 61.1 & 57.4 & 59.2 & 59.3 & 59.3 & 59.3 \\
                       & Mistral-7B  & 61.3 & 57.6 & 59.4 & 60.3 & 56.6 & 58.4 & 58.6 & 58.8 & 58.7 \\ \bottomrule
\end{tabular}
\caption{Comparison with baseline methods on Wiki-NRE. The best results are highlighted in \textbf{bold}, and the second-best results are \underline{underlined}}
\label{main_results_full_wiki}
\end{table*}

\begin{table*}[t]
\centering
\renewcommand{\arraystretch}{0.85} 
\begin{tabular}{@{}lccccccccc@{}}
\toprule
\multirow{2}{*}{F1}        & \multicolumn{3}{c}{WebNLG} & \multicolumn{3}{c}{REBEL} & \multicolumn{3}{c}{Wiki-NRE} \\ \cmidrule(lr){2-4} \cmidrule(lr){5-7} \cmidrule(l){8-10} 
          & Partial  & Strict  & Exact & Partial  & Strict & Exact & Partial  & Strict  & Exact  \\ \midrule
w.o. PO   & 72.7     & 66.7    & 69.1  & 50.0     & 44.1   & 46.3  & 62.5     & 61.2    & 61.4   \\
w.KR(cat) & 73.9     & 68.5    & 70.7  & 50.3     & 43.8   & 47.3  & 63.3     & 61.8    & 61.9   \\
w.NLI(DB) & 74.1     & 68.9    & 70.9  & 51.5     & 44.6   & 47.8  & 63.8     & 61.8    & 62.0   \\
KRPO      & \textbf{74.4}     & \textbf{69.0}    & \textbf{71.2}  & \textbf{51.9}     & \textbf{45.6}   & \textbf{48.9}  & \textbf{64.2}     & \textbf{61.8}    & \textbf{62.1}   \\ \bottomrule
\end{tabular}%
\caption{Ablation study of prompt optimizer. The best results are in bold. ``w.o.PO'' denotes removing the prompt optimization module, ``w.KR(cat)'' replaces the knowledge restoration component with naive subject-relation-object concatenation, and ``w.NLI(DB)'' replaces the llm-based NLI evaluator with the deployed DeBERTa-v3-large model.}
\label{aba_PO}
\end{table*}

\begin{table*}[t]
\centering
\renewcommand{\arraystretch}{0.85} 
\begin{tabular}{@{}lccccccccc@{}}
\toprule
\multirow{2}{*}{F1}        & \multicolumn{3}{c}{WebNLG} & \multicolumn{3}{c}{REBEL} & \multicolumn{3}{c}{Wiki-NRE} \\ \cmidrule(lr){2-4} \cmidrule(lr){5-7}\cmidrule(l){8-10} 
          & Partial  & Strict  & Exact & Partial  & Strict & Exact & Partial  & Strict  & Exact  \\ \midrule
w.o. RC & 67.8     & 60.5    & 62.8  & 46.9     & 40.3   & 43.8  & 59.3     & 57.7    & 57.9   \\
w. RC(eme)   & 73.0     & 67.4    & 69.6  & 49.5     & 43.5   & 46.8  & 62.9     & 61.4    & 61.6   \\
KRPO      & \textbf{74.4}     & \textbf{69.0}    & \textbf{71.2}  & \textbf{51.9}     & \textbf{45.6}   & \textbf{48.9}  & \textbf{64.2}     & \textbf{61.8}    & \textbf{62.1}   \\ \bottomrule
\end{tabular}%
\caption{Ablation study of relation canonicalization. The best results are in bold. ``w.o. RC'' means without the relation canonicalization module; ``w. RC(eme)'' indicates replacing the module with a cosine-similarity-based embedding matching strategy.}
\label{aba_RelCanon}
\end{table*}

\begin{table}[htbp]
    \centering
    \label{ana_tri_ele_main} 
    \begin{subtable}{0.48\textwidth}
        \centering
        \caption{Analysis of Triplet elements on WebNLG}
        \label{ana_tri_ele_webnlg}
        \begin{tabular}{@{}lccccc@{}}
            \toprule
            Method & s & r & o & (s,o) & tri \\ \midrule
            EDC    & 72.2 & 64.3 & 67.5 & 66.7 & 63.4 \\
            EDC+R  & 74.3 & 64.7 & 67.6 & 66.9 & 64.3 \\
            KRPO   & \textbf{78.2} & \textbf{70.2} & \textbf{73.7} & \textbf{72.3} & \textbf{69.0} \\ \bottomrule
        \end{tabular}
    \end{subtable}\hfill
    \begin{subtable}{0.48\textwidth}
        \centering
        \caption{Analysis of Triplet elements on REBEL}
        \label{ana_tri_ele_rebel}
        \begin{tabular}{@{}lccccc@{}}
            \toprule
            Method & s & r & o & (s,o) & tri \\ \midrule
            EDC    & 47.4 & 44.7 & 45.3 & 45.1 & 43.5 \\
            EDC+R  & 47.8 & 44.9 & 45.5 & 45.4 & 44.5 \\
            KRPO   & \textbf{49.6} & \textbf{48.5} & \textbf{48.5} & \textbf{48.2} & \textbf{45.6} \\ \bottomrule
        \end{tabular}
    \end{subtable}
        
    \begin{subtable}{0.48\textwidth}
        \centering
        \caption{Analysis of Triplet elements on Wiki-NRE}
        \label{ana_tri_ele_wikinre}
        \begin{tabular}{@{}lccccc@{}}
            \toprule
            Method & s & r & o & (s,o) & tri \\ \midrule
            EDC    & 65.1 & 61.3 & 62.9 & 61.5 & 60.6 \\
            EDC+R  & 65.3 & 61.2 & 62.9 & 61.7 & 60.8 \\
            KRPO   & \textbf{67.5} & \textbf{63.2} & \textbf{65.6} & \textbf{64.6} & \textbf{61.8} \\ \bottomrule
        \end{tabular}
    \end{subtable}\hfill
    \begin{minipage}{0.48\textwidth}
        \small
        \textbf{Note:} All metrics refer to Strict-F1 scores. Best in bold. ``s'', ``r'', ``o'', ``(s,o)'', and ``tri'' denote subject, relation, object, subject-object pair, and complete triplet.
    \end{minipage}
    \caption{Analysis of Triplet elements across datasets.}
\end{table}

\subsection{Ablation on Prompt Optimizer}
\label{appendix_ablation_PO}
To validate the effectiveness of the prompt optimizer, we conduct ablation studies by removing the prompt optimization module (w.o. PO), replacing the knowledge restoration component with naive subject-relation-object concatenation (w. KR(cat)), and substituting the LLM-based NLI evaluator with DeBERTa-v3-large (w. NLI(DB)).
As shown in Table~\ref{aba_PO}, removing the prompt optimizer leads to performance drops across all datasets, confirming its critical role. Replacing knowledge restoration with simple concatenation also degrades results, highlighting the importance of knowledge restoration for effective prompt optimization. Replacing the LLM-based NLI evaluator with DeBERTa-v3-large achieves comparable but slightly lower performance, demonstrating the advantage of LLM-based semantic feedback.

\subsection{Ablation on Relation Canonicalization}
\label{appendix_ablation_RC}
To validate the effectiveness of relation canonicalization, we conduct ablation studies by removing the relation canonicalization module (w.o. RC). As shown in Table~\ref{aba_RelCanon}, removing the relation canonicalization module leads to performance drops across all datasets, confirming its critical role. Replacing the module with a cosine-similarity-based embedding matching strategy (w. RC(eme)) also degrades results, highlighting the importance of our designed relation canonicalization for effective knowledge graph construction.

\subsection{Analysis of Triplet elements}
\label{appendix_ana_tri_ele}
To further analyze where the performance gains originate, we report Strict-F1 scores at different granularities. 
Tables~\ref{ana_tri_ele_webnlg}, \ref{ana_tri_ele_rebel} and \ref{ana_tri_ele_wikinre} shows that KRPO consistently outperforms EDC across all datasets, with the gains on the relations indicating the effectiveness of relation canonicalization. Improvements in subjects and objects suggest that prompt optimization enhances extraction stability. And the gap at the triplet level demonstrates better structural consistency under strict evaluation.

\begin{table*}[t]
\centering
\setlength{\tabcolsep}{2pt} 
\begin{tabular}{@{}lcccccccccccc@{}}
\toprule
\multirow{2}{*}{\begin{tabular}[c]{@{}l@{}}Entailment\\ Eval(Partial)\end{tabular}} & \multicolumn{4}{c}{WebNLG}                          & \multicolumn{4}{c}{REBEL}                           & \multicolumn{4}{c}{Wiki-NRE}                        \\ \cmidrule(lr){2-5} \cmidrule(lr){6-9} \cmidrule(l){10-13}
 & \multicolumn{1}{l}{Prop.} & Precision    & Recall    & F1   & \multicolumn{1}{l}{Prop.} & Precision    & Recall    & F1   & \multicolumn{1}{l}{Prop.} & Precision    & Recall    & F1   \\ \midrule
EDC    & 95\% & 67.5 & 70.5 & 68.5 & 96\% & 49.7 & 51.1 & 50.2 & 75\% & 59.2 & 60.4 & 59.7 \\
EDC+R  & 95\% & 66.7 & 70.8 & 68.7 & 96\% & 49.4 & 51.6 & 50.5 & 76\% & 60.2 & 61.6 & 60.9 \\
KRPO   & \textbf{98\%} & \textbf{72.1} & \textbf{73.7} & \textbf{72.7} & \textbf{98\%}                           & \textbf{52.0} & \textbf{53.2} & \textbf{52.4} & \textbf{94\%}                           & \textbf{62.7} & \textbf{63.4} & \textbf{63.0} \\ \bottomrule
\end{tabular}%
\caption{Entailment Triplet Analysis. Best results are in bold. Prop. is the entailment proportion by self-evaluation.}
\label{tab:entail_detail}
\end{table*}

\subsection{Performance analysis of self-evaluation mechanism in KRPO}
\label{appendix:entailment}
To further validate the effectiveness of our proposed self-evaluation mechanism, we conduct an in-depth analysis of the entailment triplets generated during the extraction process.
As shown in Table~\ref{tab:entail_detail}, KRPO achieves a higher entailment proportion across all datasets compared to EDC. This indicates that the triplets generated by KRPO are more semantically consistent with the source text, demonstrating the effectiveness of our self-evaluation strategy in enhancing the quality of extracted information.

\subsection{Analysis of Token Consumption}
\label{sec:token_analysis}

\begin{figure*}[t]
    \centering
    \includegraphics[width=0.8\textwidth]{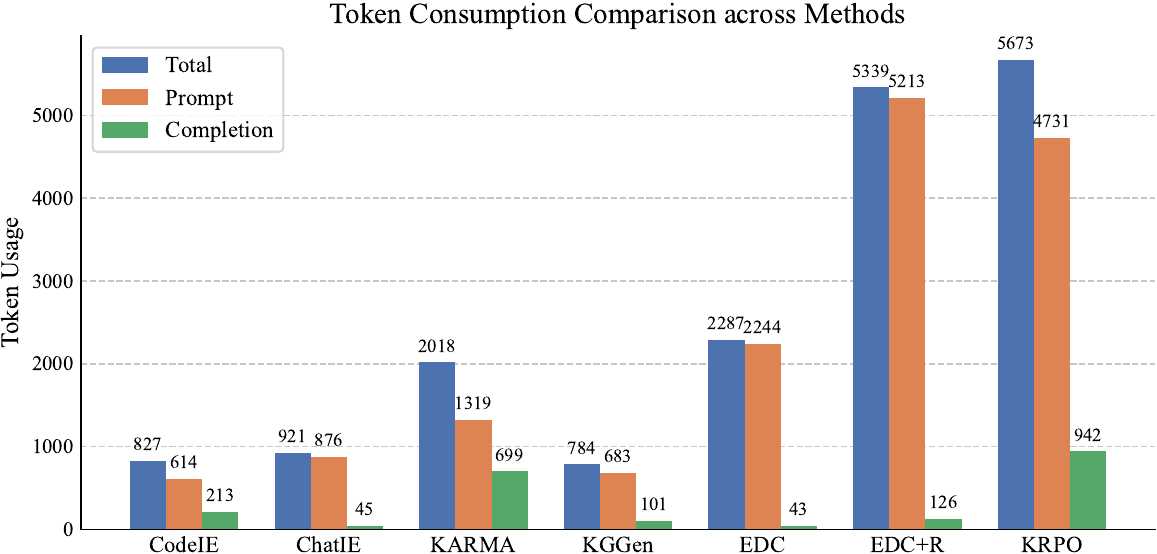}
    \caption{Analysis of Token Consumption.}
    \label{fig:token_cost}
\end{figure*}

Figure~\ref{fig:token_cost} breaks down the token consumption of different methods into prompt (input) and completion (output) tokens, evaluated on the average consumption of 25 test samples using the GPT-4o-mini backbone. Overall, KRPO incurs the highest total token usage, slightly exceeding the strongest baseline, EDC+R. Early LLM or Agent-based methods like CodeIE and ChatIE are much more token-efficient, but as shown in our main results, they perform significantly worse in open-domain extraction.

When separating the total cost, we observe a clear difference in how KRPO and EDC+R consume resources. Interestingly, KRPO actually uses fewer prompt tokens compared to EDC+R. While EDC+R attempts to maintain performance by packing the context window with extensive static schema descriptions and demonstrations, our method distills the task instructions into a more efficient, dynamically optimized prompt, effectively reducing the input burden.
The primary source of KRPO's overhead lies in the completion phase, which consumes noticeably more tokens than static methods. This increase is a direct result of the proposed feedback loop. In essence, KRPO trades higher output generation costs for the ability to self-reflect and self-correct. Considering the substantial improvements in strict extraction metrics, we view this as a reasonable trade-off for applications demanding high-quality knowledge graphs.

\subsection{Analysis of Extraction Density}
Table~\ref{tab:avg_triplet_num} reports the average number of triplets extracted per sample, with the values in parentheses indicating deviations from the reference annotations. On WebNLG, both EDC+R and KRPO produce extraction densities close to the reference value of 3.43, while KRPO remains particularly well aligned across different LLM backbones. On REBEL and Wiki-NRE, KRPO consistently extracts more triplets than EDC+R, suggesting that prompt adaptation encourages the models to identify a broader set of relational information. The increase is most evident on REBEL, where KRPO exceeds EDC+R by approximately 0.21--1.27 triplets per sample across the three backbones.

A larger number of extracted triplets does not necessarily indicate better extraction, as it may also introduce unsupported relations. However, when considered together with the F1 improvements reported in the main results, these observations suggest that KRPO improves relation coverage rather than merely promoting uncontrolled over-extraction. The relatively large deviations on Wiki-NRE also indicate that both methods tend to generate more relations than the reference annotations, highlighting the importance of jointly considering extraction density, precision, and recall.

\begin{table*}[]
\centering
\begin{tabular}{@{}clccc@{}}
\toprule
\multicolumn{1}{c}{Methods}&LLM for ORTE             & WebNLG       & REBEL          & Wiki-NRE     \\ \midrule
\multicolumn{1}{c}{Reference} &-                 & 3.43         & 4              & 2.86         \\ \midrule
\multicolumn{1}{l}{REGEN/GENIE} & -            & -            & 2.2(-1.80)     & 3.08(+0.22)  \\ \midrule
\multirow{3}{*}{EDC+R}          & DeepSeekV3   & 3.57 (+0.14) & 4.46 (+0.46)   & 4.00 (+1.14) \\
                                & GPT-4o-mini  & 3.09 (-0.34) & 4.40 (+0.40)   & 4.02 (+1.16) \\
                                & Mistral7b-v3 & 3.48 (+0.05) & 4.68 (+0.68)   & 3.96 (+1.10) \\ \midrule
\multirow{3}{*}{KRPO}           & DeepSeekV3   & 3.72 (+0.29) & 5.41 (+1.41)   & 4.37 (+1.51) \\
                                & GPT-4o-mini  & 3.39 (-0.04) & 5.67 (+1.67)   & 4.37 (+1.51) \\
                                & Mistral7b-v3 & 3.50 (+0.07) & 4.893 (+0.893) & 4.19 (+1.33) \\ \bottomrule
\end{tabular}%
\caption{Average number of extracted triplets per sample across different ORTE methods and LLM backbones. Values in parentheses denote the difference from the average number of reference triplets.}
\label{tab:avg_triplet_num}
\end{table*}

\section{Case study of detailed KRPO process}
\label{appendix_case_study}
To show our algorithm flow in more detail, we use an example in this chapter to go step-by-step through the method details.

\textbf{ORTE.} Before the task starts, we need an initial Prompt to initialize ORTE. The initial prompt $p$ is as follows:

\begin{tcolorbox}[colback=white!5!white, colframe=gray!18!gray, title=$p$: Initial Prompt for ORTE]
Your task is to transform the given text into a semantic graph in the form of a comprehensive list of triplets.\\
The triplets must be in the form of [Entity1, Relationship, Entity2].\\
In your answer, please strictly only include the triplets list and do not include any explanation or apologies.
\end{tcolorbox}

Given the input sample sentence: ``\textit{In the city of Detroit in the state of Michigan, the first Pontiac Rageous was produced on the assembly line in 1997.}'', we initially used the initial prompt $p$ to extract the following relation triplets:
\begin{itemize}
    \item \textit{[Pontiac Rageous, assembly, Detroit]}
    \item \textit{[Pontiac Rageous, buildDate, 1997]}
    \item \textit{[Pontiac Rageous, state, Michigan]}
\end{itemize}
\textbf{Self Evaluation.} 
In the second stage, we applied the prompt for Knowledge Restoration to guide the LLM in reconstructing these triplets into natural language. The reconstructed sentences are as follows:
\begin{itemize}
    \item ``\textit{The Pontiac Rageous was assembled in Detroit.}''
    \item ``\textit{The Pontiac Rageous was built in 1997.}''
    \item ``\textit{The Pontiac Rageous is associated with the state of Michigan.}''
\end{itemize}

Next, we evaluated the semantic consistency between the reconstructed sentences and the original input text using the NLI evaluation prompt. Detailed evaluation results are as follows:

\begin{quote}
\{\\
\hspace*{1em}``\textit{label}'': ``\textit{entailment}'',\\
\hspace*{1em}``\textit{confidence}'': 0.95,\\
\hspace*{1em}``\textit{reasoning}'': ``\textit{The premise explicitly states that the first Pontiac Rageous was produced on the assembly line in Detroit, thus confirming that the Pontiac Rageous was assembled in Detroit.}''\\
\}
\end{quote}

\begin{quote}
\{\\
\hspace*{1em}``\textit{label}'': ``\textit{entailment}'',\\
\hspace*{1em}``\textit{confidence}'': 0.9,\\
\hspace*{1em}``\textit{reasoning}'': ``\textit{The premise states that the first Pontiac Rageous was produced on the assembly line in 1997 in Detroit. This supports the hypothesis that the Pontiac Rageous was built in 1997, making the hypothesis logically follow from the premise.}''\\
\}
\end{quote}

\begin{quote}
\{\\
\hspace*{1em}``\textit{label}'': ``\textit{neutral}'',\\
\hspace*{1em}``\textit{confidence}'': 0.85,\\
\hspace*{1em}``\textit{reasoning}'': ``\textit{The premise states that the Pontiac Rageous was produced in Detroit, Michigan, in 1997, but it does not provide information about the current location of the Pontiac Rageous. Hence, the hypothesis that asks if the Pontiac Rageous is currently in Michigan is neither confirmed nor denied by the premise.}''\\
\}
\end{quote}

\textbf{Prompt Optimization.} 
Next, we perform text gradient-based prompt optimization based on these evaluation results.
We apply the instructions $I_1$ and $I_2$ in the optimizer to each triplet separately.
Taking these triplets, we obtain the evaluation-driven feedback on ``\textit{how to improve evaluation indicators}'' by $I_1$ as follows:

\begin{quote}
``\textit{
Correctness:\\
Ensure that all elements of the triplet are accurately derived from the source text. In this case, verify if the relationship `state' truly describes the connection between `Pontiac Rageous' and `Michigan.'
If the extracted term is not reflective of the actual relation depicted in the source, it may require re-evaluation. Consider revising the relation to something more precise if applicable.\\
Completeness:\\
Ensure that all relevant context is captured within the triplet. If `Detroit, Michigan' is the specific location mentioned, it might be beneficial to capture the city for more precision.\\
Clarity:\\
Clearly define the nature of the relation so that it is unambiguous. If `state' lacks clarity, replace it with a more explicit term that reflects the nature of the relationship.\\
Ensure that end-users understand what the relationship describes. If the triplet is part of a larger dataset, maintain consistent terminology across entries.\\
Re-evaluation of the Context:\\
Review the source context to verify that the extracted triplet holds true under likely interpretations. If the premise doesn't support the hypothesis concretely, consider rephrasing the triplet to match the available evidence in the text.
}''
\end{quote}

Based on this feedback, we obtained the following prompt optimization guidance based on the prompts obtained by $I_2$:
\begin{quote}
``\textit{
To enhance the system prompt for Relational Triplet Extraction, several targeted modifications are necessary:\\
Clarify Relationship Extraction: Encourage accuracy by specifying that the relationship term must directly reflect the semantic link in the context. Update the prompt to include examples of explicit relationships to guide the model toward correct interpretations, especially when multiple relationships might be inferred.\\
Example: `Transform the given text into a semantic graph of triplets, with relationships reflecting direct context links'\\
Emphasize Completeness: Instruct the model to prioritize capturing all contextually relevant details, ensuring precise location-based relationships are extracted. Add guidelines for interpreting nested or compound locations like `city, state' instead of only considering individual elements.\\
Example: `Ensure completeness by evaluating compound locations and specific contexts thoroughly, e.g., `Detroit, Michigan'.\\
Enhance Clarity: Define the expected structure and context of the triplets more clearly. Suggest using clear and standardized relationship terminology within the dataset's context, ensuring reliable consistency.\\
Example: `Maintain clarity and consistency in relationships, preferring standard terms within the dataset's context.'\\
Re-evaluate Contextually: Encourage a thorough re-evaluation of the text to confirm that extracted relationships are contextually supported, preventing assumptions not rooted in the provided information.
}''
\end{quote}

After obtaining the prompt guidance strategies for all samples in a batch, we integrate them into $I_3$ and guide the LLM to update the prompts. The updated prompt is as follows:

\begin{tcolorbox}[colback=white!5!white, colframe=gray!18!gray, title=$p^*$: Updated Prompt for ORTE]
Your task is to transform the given text into a semantic graph represented as a list of relational triplets. \\
- Extract all explicitly stated, text-supported relationships between entities, without omitting any valid relations mentioned in the input.\\
- Ensure that each extracted triplet captures a single, atomic, and contextually accurate relation, using concise and unambiguous relationship phrases faithful to the semantics expressed in the text. \\
- Avoid inferred, explanatory, or implicit relations that are not explicitly stated.\\
- Present the triplets in the format [Entity1, Relationship (clear term), Entity2]. \\
In your answer, output only the list of triplets and do not include any additional text.
\end{tcolorbox}

\textbf{ORTE after Prompt Optimization.} 
Based on the optimized prompts, we guide LLM to perform ORTE again, and the newly extracted triplets are as follows:
\begin{itemize}
    \item \textit{[Detroit, state, Michigan]}
    \item \textit{[Pontiac Rageous, productionYear, 1997]}
    \item \textit{[Pontiac Rageous, productionLocation, Detroit]}
    \item \textit{[Pontiac Rageous, productionLocation, Michigan]}
\end{itemize}

\textbf{Self Evaluation after Prompt Optimization.}
We again apply the prompt for Knowledge Restoration to guide the LLM in reconstructing these new triplets into natural language and evaluate their semantic consistency with the original input text using the NLI evaluation prompt. The evaluation results corresponding to the four extracted triplets above are all ``\textit{entailment}'' with high confidence scores.

\textbf{Relation Canonicalization.}
Finally, we perform relation canonicalization on the extracted triplets. By calculating the semantic similarity between the extracted relations and schemas in memory, we map them to their canonical forms or update the memory. For instance, ``\textit{productionYear}'' is mapped to ``\textit{buildDate}'', and ``\textit{productionLocation}'' is added in memory as a new relation. The final canonical triplets are as follows:
\begin{itemize}
    \item \textit{[Detroit, state, Michigan]}
    \item \textit{[Pontiac Rageous, buildDate, 1997]}
    \item \textit{[Pontiac Rageous, productionLocation, Detroit]}
    \item \textit{[Pontiac Rageous, productionLocation, Michigan]}
\end{itemize}

Ultimately, these triplets are integrated into the knowledge graph.

\section{Discussion on Evaluation Metrics for Open-Domain RTE}
\label{appendix:metric_discussion}

Current evaluation protocols for Relational Triplet Extraction (RTE) predominantly rely on surface-level matching, evaluating the correctness of extracted entities and relations based on exact string match or character/token-level overlap. While these metrics are well-suited for closed-domain extraction with predefined schemas, they exhibit significant limitations in open-domain settings.

In Open Relational Triplet Extraction (ORTE), relations are schema-extensible and highly diverse. LLM-based generators frequently produce relations that are semantically accurate but lexically distinct from the golden annotations. For instance, extracting ``\textit{was founded by}'' instead of the annotated ``\textit{creator}'', or ``\textit{is located in}'' instead of ``\textit{country}''. Traditional lexical metrics rigidly penalize these valid extractions as false negatives. Consequently, these metrics fail to reflect the true semantic fidelity of the extracted knowledge and often severely underestimate the capabilities of generative frameworks.

This discrepancy underscores an urgent need for the community to transition toward deep, semantic-level evaluation paradigms. Future evaluation frameworks should incorporate semantic similarity scoring, such as Cross-Encoder verification, dense embedding alignment, or LLM-as-a-judge mechanisms to evaluate the underlying meaning of triplets rather than their literal manifestations. In this work, our Relation Canonicalization module serves as a crucial intermediate step to align free-form generations with standard schemas, effectively mitigating the misalignment under current rigid metrics. However, developing robust, standardized semantic evaluation protocols remains a critical future direction for advancing open-domain knowledge graph construction.


\end{document}